\documentclass{siamonline1116}

\usepackage{lipsum}
\usepackage{amsfonts}
\usepackage{graphicx}
\usepackage{epstopdf}
\usepackage{algorithmic}
\ifpdf
  \DeclareGraphicsExtensions{.eps,.pdf,.png,.jpg}
\else
  \DeclareGraphicsExtensions{.eps}
\fi

\numberwithin{theorem}{section}

\newcommand{\TheTitle}{Non-negative Tensor Patch Dictionary Approaches for Image Compression and Deblurring Applications} 
\newcommand{\TheAuthors}{Elizabeth Newman and Misha E. Kilmer}

\headers{Tensor Patch Dictionaries}{\TheAuthors}
\title{{\TheTitle}\thanks{Submitted to the editors DATE.
\funding{This research is partially based upon work supported by the National Science Foundation under NSF 1319653 and by the Office of the Director of National Intelligence (ODNI), Intelligence Advanced Research Projects Activity (IARPA), via IARPA’s 2014-14071600011. The views and conclusions contained herein are those of the authors and should not be interpreted as necessarily representing the official policies or endorsements, either expressed or implied, of ODNI, IARPA, or the U.S. Government. The U.S. Government is authorized to reproduce and distribute reprints for Governmental purpose notwithstanding any copyright annotation thereon. }}}

\author{Elizabeth Newman\thanks{Department of Mathematics, Emory University, Atlanta, GA
    (\email{elizabeth.newman@emory.edu})}
  \and
  Misha E. Kilmer\thanks{Department of Mathematics, Tufts University, Medford, MA
   (\email{misha.kilmer@tufts.edu}, \url{http://www.tufts.edu/\~mkilme01/}).}
}

\usepackage{amsopn}

\usepackage{subfigure} 

\usepackage{float} 

\usepackage{amsmath}
\usepackage{mathtools}

\usepackage{tikz}

\usepackage{arydshln} 

\usepackage{float} 

\usepackage{bbm}

\usepackage{bm}

\usepackage[mathscr]{eucal}

\newtheorem{example}{Example}[section]

\newcommand{\bs}{\boldsymbol}
\DeclareMathOperator*{\argmin}{arg\,min}

\newcommand{\unfold}{\texttt{unfold}}
\newcommand{\fold}{\texttt{fold}}

\newcommand{\Rbb}{\mathbb{R}}
\newcommand{\V}[1]{{\bm{\mathbf{\MakeLowercase{#1}}}}} 

\newcommand{\M}[1]{{\bm{\mathbf{\MakeUppercase{#1}}}}} 

\newcommand{\T}[1]{\boldsymbol{\mathscr{\MakeUppercase{#1}}}} 

\newcommand{\TA}{\T{A}}
\newcommand{\TB}{\T{B}}
\newcommand{\TC}{\T{C}}
\newcommand{\TX}{\T{X}}
\newcommand{\TU}{\T{U}}
\newcommand{\TV}{\T{V}}
\newcommand{\TS}{\T{S}}

\newcommand{\TD}{\T{D}}

\newcommand{\MZ}{\M{Z}}

\newcommand{\MA}{\M{A}}
\newcommand{\MX}{\M{X}}

\newcommand{\MD}{\M{D}}
\newcommand{\MB}{\M{B}}
\newcommand{\MP}{\M{P}}
\newcommand{\MQ}{\M{Q}}

\newcommand{\MI}{\M{I}}
\newcommand{\ML}{\M{L}}

\newcommand{\bea}{\left[ \begin{array}} 
\newcommand{\eea}{\end{array} \right] }

\ifpdf
\hypersetup{
  pdftitle={\TheTitle},
  pdfauthor={\TheAuthors}
}
\fi

\newcommand{\Circ}[1]{{\tt circ}\left( #1 \right)}
\newcommand{\Fold}[1]{{\tt fold}\left( #1 \right)}
\newcommand{\Unfold}[1]{{\tt unfold}\left( #1 \right)}

\newcommand{\Squeeze}[1]{{\tt squeeze} \left( #1 \right)} 

\newcommand{\Trace}[1]{{\tt trace}\left( #1 \right)}

\begin{document}

\maketitle

\begin{abstract}
 In recent work (Soltani, Kilmer, Hansen, BIT 2016), an algorithm for non-negative tensor patch dictionary learning in the
 context of X-ray CT imaging and
 based on a tensor-tensor product called the $t$-product (Kilmer and Martin, 2011) was presented.  
Building on that work, in this paper, we use of non-negative tensor patch-based dictionaries trained on other data, such as facial image data, for the purposes of either compression or image deblurring.    
We begin with an analysis in which we address issues such as suitability of the tensor-based approach relative to a matrix-based approach, dictionary size and patch size to balance computational efficiency and qualitative representations. 
Next, we develop an algorithm that is capable of recovering non-negative tensor coefficients given a non-negative tensor dictionary.  The algorithm is based on a variant of the Modified Residual Norm Steepest Descent method.  We show how to augment the algorithm to enforce sparsity in the tensor coefficients, and note that the approach has broader applicability since it can be applied to the matrix case as well. 
We illustrate the surprising result that dictionaries trained on image data from one class can be successfully used to represent and compress image data from different classes and across different resolutions.  
Finally, we address the use of non-negative tensor dictionaries in image deblurring.  We show that tensor treatment of the deblurring problem coupled with non-negative tensor patch dictionaries can give superior restorations as compared to 
standard treatment of the non-negativity constrained deblurring problem.   
 
\end{abstract}



\section{Introduction}
\label{sec:introduction}

	Many applications in imaging science, such as image deblurring and image reconstruction, typically model the object to be recovered as a vector of unknowns, and the forward operator as a matrix.   
Treating image and video data processing problems using tensor approaches is yet a new, but increasingly popular and promising approach, as suggested by recent literature \cite{HaoEtAl2013,Martin:2013kx, liu2013tensor, hunyadi2017tensor, vasilescu2002multilinear, semerci2013tensor,NewKilHor17,Soltani2016,ZZhangetal}.   
However, a close look at the increasing body of literature in which tensor decompositions are used in practice shows that no one specific tensor decomposition has fit all these image and video applications equally well.    Indeed, here, as in many other multiway data processing and representation problems, the type of decomposition to be employed may be quite specific to the application.   

Decompositions based on a tensor-tensor product called the $t$-product \cite{KilmerMartin2011} have proven to be particularly useful in applications where there is a natural orientation dependence to be preserved, such as pixel or voxel position, relative to some other variable such as number of images or time  (see \cite{HaoEtAl2013,ZZhangetal} for example).   The $t$-product is advantageous over other tensor decompositions because of the algebraic framework induced by the definition of the $t$-product, which enables the definition and computation of factorizations reminiscent of their matrix counterparts (e.g. SVD, QR) and because the products can be computed in a straightforward way in parallel (see also \cite{KilmerEtAl2013}).  
    
Non-negative tensor factorizations have been introduced in the literature recently as well, and just like the unconstrained counterparts, the type of decomposition used varies \cite{Cickbook}.   In \cite{Soltani2016, HaoEtAl2013}, the authors consider non-negative tensor decompositions based specifically on the $t$-product, the approaches in the two papers differing by the additional constraints on the optimization as well as the algorithms proposed to compute the factorization.    In \cite{Soltani2016}, the authors develop an Alternating Direction Method of Multipliers (ADMM) \cite{Boyd2010} method for producing a non-negative patch tensor dictionary from a single, high resolution training image with the end goal of using the dictionary in the context of X-ray CT image reconstruction.  The authors showed the method was good at producing reconstructions even for missing data situations, and that it gave improvements over matrix-based patch dictionary learning since the reconstructions were sparser and less sensitive to regularization parameters. 

In this paper, we consider two classical problems -- (lossy) image compression and image deblurring.  In both applications, the first stage is to learn a tensor patch dictionary from multiple images of the same class using the approach in \cite{Soltani2016}, and hence we review that problem briefly.  Our first new contribution deals with finding a non-negative representation under the tensor $t$-product \cite{KilmerMartin2011} of any image given a tensor dictionary.  We give theoretical results and concrete illustrations that demonstrate the superiority of a tensor-patch dictionary over the corresponding matrix case. 
Then, we show how the modified residual norm steepest descent (MRNSD), \cite{Nagy2000,Kaufman1993} can be utilized for non-negative tensor coefficient recovery under the $t$-product.  Additionally, we introduce sparsity-inducing regularization to the  algorithm that can lead to compressed representations for images.   Furthermore, we show that constraining our image to the non-negative patch dictionary representation can lead to a new effective debluring approach that is robust to 
certain model mismatches.

 This paper is organized as follows.   \Cref{sec:back} is devoted to the introduction of background and notation.      In \Cref{sec:patches}, we describe the process of patchification of images to make a tensor representation, and review the dictionary learning approach presented in \cite{Soltani2016} which we will use to generate our tensor dictionaries.   In \Cref{sec:superior}, we investigate the power of the tensor-tensor product based representation of images vs. the traditional matrix-based approach.  
The choice of parameters such as patch and dictionary sizes are relative to quality, storage and
computation time are also considered here. 
Following that discussion is the MRNSD algorithm for tensors in \Cref{sec:mrnsd}.  Here, we also discuss the incorporation of coefficient sparsity constraints into MRNSD to allow for the compressed representation of images.    In \Cref{sec:deblur} we give a short introduction to the image deblurring problem, explain how to represent the unknown image in terms of the tensor dictionary, and discuss the restoration problem that needs to be solved for the tensor coefficients.   Numerical results are contained in Section \Cref{sec:num} and a discussion and list of future work is given in \Cref{sec:future}.  Detailed derivations for some of the claims are left to the appendicies.  

\section{Notation and preliminaries}

\label{sec:back}

A \emph{tensor} is a multidimensional array of data; a first-order tensor is a vector and a second-order tensor is a matrix.  This paper focuses on third-order tensors (i.e., three-dimensional data), though much of the theory can be extended to higher-order tensors.  We denote tensors with script letters.

\begin{figure}[H]
	\centering
	\subfigure[Tensor $\TA$]{\includegraphics[scale=0.25]{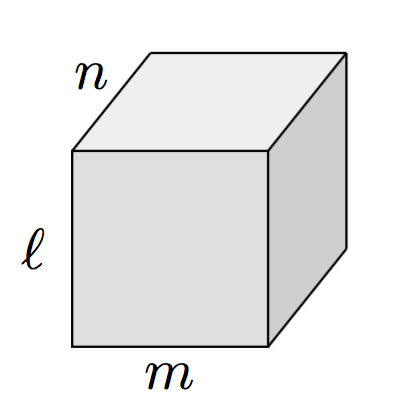}}
	\subfigure[Frontal slices $\MA^{(k)}$]{\includegraphics[scale=0.25]{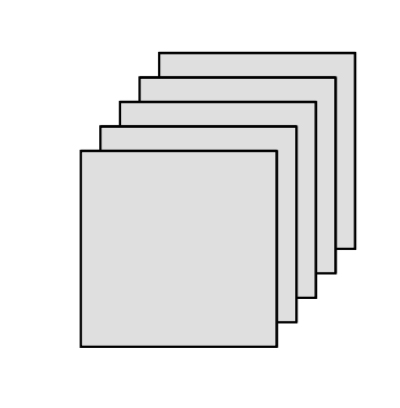}}
	\subfigure[Lateral slices $\vec{\TA}_j$]{\includegraphics[scale=0.25]{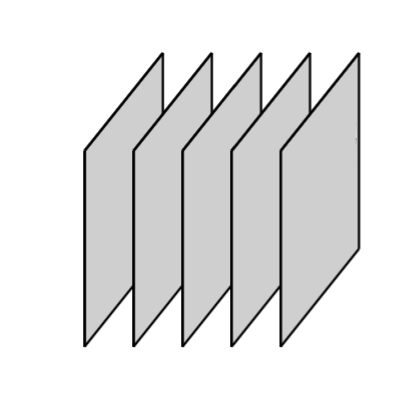}}
	\subfigure[Tubes $\V{a}_{ij}$]{\includegraphics[scale=0.25]{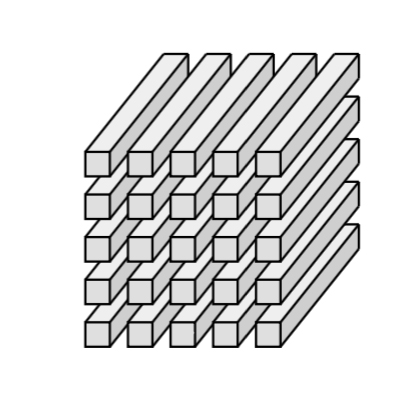}}
	\caption{Tensor notation.}
	\label{fig:tensor notation}
	\end{figure}

Suppose $\TA$ is an $\ell \times m\times n$ tensor.  As depicted in \Cref{fig:tensor notation}, we can divide the tensor in several directions.  \emph{Frontal slices}, denoted $\MA^{(k)}$ for $k = 1,\dots, n$,  are $\ell\times m$ matrices which fix the third-dimension of $\TA$.  \emph{Lateral slices}, denoted $\vec{\TA}_j$ for $j = 1,\dots, m$, are $\ell\times 1\times n$ tensors which fix the second-dimension of $\TA$; we consider lateral slices to be $\ell\times n$ matrices oriented along the third-dimension.  \emph{Tube fibers or tubes}, denote $\V{a}_{ij}$ for $i = 1,\dots, \ell$ and $j = 1,\dots, m$, are the $1\times 1\times n$ mode-3 fibers of $\TA$ or $n \times 1$ column vectors oriented along the third dimension.

If we consider tensors as linear operators analogous to matrices, lateral slices are the analogous to column vectors, hence the notation $\vec{\TA}$.  In particular, tensors act on lateral slices just as matrices act on column vectors.  Furthermore, lateral slices form the range and null space of a tensor \cite{KilmerEtAl2013}.  For more detailed analysis on tensor linear algebra, we reference \cite{KilmerEtAl2013}.
 
Many of the following definitions are taken directly from \cite{KilmerMartin2011}.
Using the $\ell\times m\times n$ tensor $\TA$ illustrated in \Cref{fig:tensor notation}, we define the \unfold\ and \fold\ operations as follows:
	\begin{equation}\label{eqn:unfold}
	\unfold(\TA) = \underset{\begin{array}{c}\\[-1em] \ell n\times m\end{array}}{\begin{pmatrix}\MA^{(1)} \\ \MA^{(2)}\\ \vdots \\ \MA^{(n)}\end{pmatrix}}, \qquad
	\fold(\unfold(\TA)) = \TA.
	\end{equation}
The \unfold\ function reshapes a tensor $\TA$ into a block-column vector where each block is a frontal slice.  The \fold\ function reshapes an unfolded tensor into its original structure.  Notice that the number of elements of $\TA$ and $\unfold(\TA)$ is the same. 

We now define the function {\tt circ} which transforms a tensor $\TA$ into a block-circulant matrix whose blocks are the frontal slices of $\TA$.
	\begin{equation}\label{eqn:bcirc}
	\Circ{\TA} = \underset{\begin{array}{c}\\[-1em] \ell n\times mn\end{array}}
			{\begin{pmatrix}
			\MA^{(1)} & \MA^{(n)} & \dots & \MA^{(2)}\\
			\MA^{(2)} & \MA^{(1)} & \dots & \MA^{(3)}\\
			\vdots  & \vdots  & \ddots & \vdots\\	
			\MA^{(n)} & \MA^{(n-1)} & \dots & \MA^{(1)}\\
			\end{pmatrix}}.
	\end{equation}
 Notice that the first column of $\Circ{\TA}$ is the unfolded tensor from \Cref{eqn:unfold}.  Furthermore, notice that $\Circ{\TA}$  has $n$ times the number of elements of the original tensor $\TA$.  Fortunately, we need not form $\Circ{\TA}$ explicitly.

 Using \cref{eqn:unfold} and \cref{eqn:bcirc}, the $t$-product of two tensors is defined in \cite{KilmerMartin2011} as follows:
 	\begin{definition}[$t$-product]\label{defn:tprod}
	Given $\TA$ is an $\ell\times p \times n$ tensor and $\TB$ is $p\times m\times n$, we define the \emph{$t$-product} as
		\begin{align*}
		\TA * \TB = \Fold{\Circ{\TA} \cdot \Unfold{\TB}},
		\end{align*}
	where $\TA * \TB$ is an $\ell\times m\times n$ tensor and $``*"$ denote the $t$-product.
	\end{definition}
Note that tubes commute under the $t$-product, and thus act analogously to scalars.

For the algorithms we describe in \Cref{sec:mrnsd}, we require the following two tensor norms \cite{KoldaBader,Soltani2016}.
	\begin{definition}[Frobenius norm]\label{defn:fronorm}
	Suppose $\TA$ is an $\ell\times m\times n$ tensor.  Then:
		\begin{align*}
		\|\TA \|_F^2 = \Trace{(\TA^T*\TA)^{(1)}} 
			= \sum_{k=1}^n\sum_{j=1}^m\sum_{i=1}^\ell (\TA_{ij}^{(k)})^2.
		\end{align*}
	\end{definition}

	\begin{definition}[Sum norm]\label{defn:sum norm}
	Suppose $\TA$ is an $\ell\times m\times n$ tensor.  The sum norm is
		\begin{align*}
		\|\TA \|_{\textnormal{sum}} = \sum_{k=1}^n\sum_{j=1}^m\sum_{i=1}^\ell |\TA_{ij}^{(k)}|.
		\end{align*}
	\end{definition}

\subsection{Properties of the $t$-product}
\label{ssec:props}

As we alluded to above and given in \cite{KilmerMartin2011}, we can compute the $t$-product (see \Cref{defn:tprod}) more efficiently using the Fourier transform: 
	\begin{definition}[$t$-product with Fourier transform]\label{defn:tprod fourier}
	Given $\TA$ is an $\ell\times p \times n$ tensor and $\TB$ is $p\times m\times n$, the $t$-product $\TC = \TA*\TB$ can be computed as follows:
	\begin{align*}
	\widehat{\M{C}}^{(i)} = \widehat{\MA}^{(i)} \cdot \widehat{\MB}^{(i)} \quad \text{for }i = 1,\dots n,
	\end{align*}
	where $\widehat{\TA} = \mbox{\tt fft}(\TA,[\,],3)$, $\TC = \mbox{\tt ifft}(\widehat{\TC},[\,],3)$, and \mbox{\tt fft}, \mbox{\tt ifft} are the one-dimensional fast Fourier and inverse Fourier transforms, respectively, applied along the third-dimension.
	\end{definition}

\Cref{defn:tprod fourier} can be implemented in parallel perfectly, hence is an efficient algorithm for computing the $t$-product.
	
\begin{figure}[H]
	\centering
	\includegraphics[scale = 0.35]{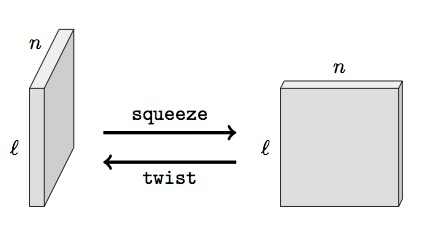}
	\caption{Illustration of \mbox{\tt squeeze} and \mbox{\tt twist} transformations.}
	\label{fig:squeeze}
	\end{figure}

	An alternative perspective on the $t$-product will be essential to our understanding of tensor dictionary learning in \Cref{subsec:localglobal}. (See also \cite{HaoEtAl2013,Soltani2016}.)
 Suppose $\vec{\TA}$ is an $\ell\times 1\times n$ lateral slice.  Then, $\mbox{\tt squeeze}(\vec{\TA})$ rotates the lateral slice into an $\ell\times m$ matrix; the \texttt{twist} transformation reverses this process (see \Cref{fig:squeeze}).
	
Next we show how the structure imposed by the $t$-product impacts lateral slices.

	\begin{definition}[$t$-product with lateral slices] \label{defn:tprodlat}
	Given $\TA$ is an $\ell\times p \times n$ tensor and $\TB$ is $p\times m\times n$, we can write the $\TC = \TA*\TB$ as follows:
		\begin{align*}
		\Squeeze{\vec{\TC}_j} = \sum_{i=1}^p \Squeeze{\vec{\TA}_i} \cdot \mbox{\tt circ}(\V{b}_{ij}^T)
		\quad \text{for } j = 1,\dots, m.
		\end{align*}
	\end{definition}

Originally, we viewed the $t$-product as $\TA$ acting on the lateral slices of $\TB$ (see \Cref{defn:tprod}). The significance of \Cref{defn:tprodlat}, originally noted in \cite{HaoEtAl2013}, is that we can consider tubes of $\TB$ to be ``coefficients" of lateral slices of $\TA$.   We say more about this in Section \ref{subsec:localglobal}.

\section{Patch Tensor Representation and Learning}
\label{sec:patches}

We briefly discuss the general idea of dictionary learning with tensors.  For more background on matrix-based dictionary learning, one can see \cite{SoltAndHans16} and the references therein.  As this paper focuses on tensor formulations, we keep our overview of the literature to describing the tools from \cite{Soltani2016} that we use here.
	

\subsection{Image to Tensor Mapping}  \label{ssec:patch}

\begin{figure}[htb]
 \centering
\includegraphics[scale=.35]{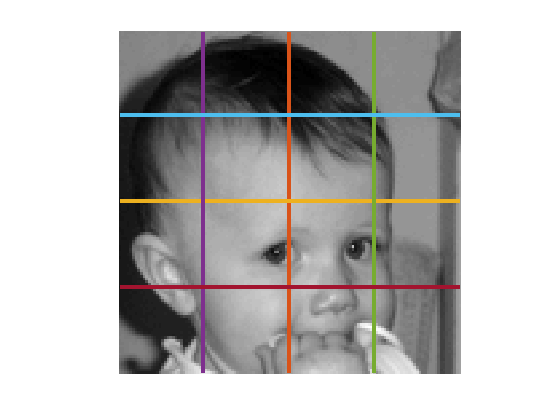} \includegraphics[scale=.35]{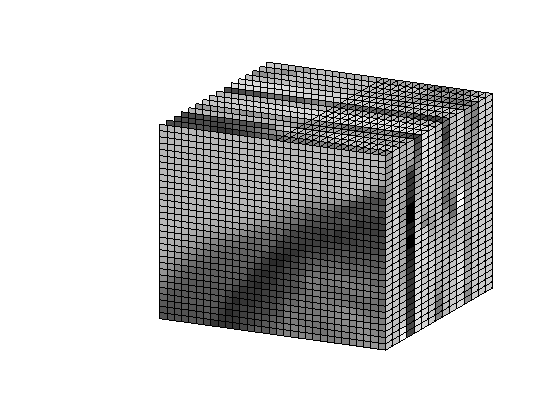} 
\caption{Illustration of tensor patchification of image (left) to construct its tensor representation (right).}
\label{fig:patchifying}
\end{figure}

First, let us describe the transformation of a single two dimensional image into a third-order tensor.  Let us suppose we have one image $\MB$ of size $N_r \times N_c$, and we desire to consider this image in terms of $p \times q$ patches, where $N_r = p n_r$ and $N_c = q n_c$ for some integers $n_r, n_c$, respectively.  Then our $N_r \times N_c$ image can be mapped to a $p \times n_r n_c \times q$ third order tensor $\TB$ by putting each image patch into a lateral slice of our tensor.  We choose to use a lexicographical ordering by patch columns.   As seen in \Cref{fig:patchifying}, the (1,1) patch in $\MB$ is mapped to the first lateral slice in $\TB$ (i.e. $\TB_{:,1,:}$), the (2,1) patch in $B$ becomes the 2nd lateral slice of $\TB$, etc.   Clearly, the process is completely reversible:  given the patch tensor representation of an image, we can map back to its matrix representation.      

In sum, $\V{b}, \MB, \TB$ all represent the same image, but in different formats.  Relative dimensions are summarized in
\Cref{tab:sizes} for easy reference.

\begin{table}[H]
\centering
\begin{tabular}{|c|c|c|c|c|} \hline
\multicolumn{5}{|c|}{$N_r = n_r p; \mbox{  } N_c = n_c q; \mbox{  } M = n_r n_c$} \\ \hline
$\MB$ & $\TB$ & $\V{b}$ & $\TD$ & $\TC$ \\ \hline
$N_r \times N_c$ & $p \times M \times q$ & $N_r N_c \times 1 $ & $p \times s \times q$ & $s \times M \times q$ \\ \hline
\end{tabular}
\caption{\label{tab:sizes} Left three columns give the dimensions of the various representations of the same image.  In the right two columns, the image approximation in tensor form is assumed $\TX = \TD * \TC$, and the corresponding sizes of $\TD$ and $\TC$ under this assumption are given. }
\end{table}

\subsection{Tensor-based dictionary learning}
\label{ssec:tendict}

\begin{figure}[H]
	\centering
	\includegraphics[scale = 0.4]{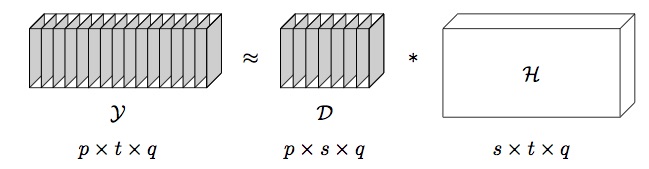}
	\caption{Illustration of tensor dictionary learning decomposition.}
	\label{fig:tensor dictionary}
	\end{figure}

	Now suppose we have a sample space of $N_I$ images, each image of size $N_r \times N_c$.  Following \cite{Soltani2016}, we divide each image into patches of size $p\times q$ and let $M$ be the number of patches per image.  Unlike matrix-based dictionary learning, we do not vectorize each patch.  Instead, we store all patches as lateral slices of a sample space tensor $\T{Y}$ of size $p\times t\times q$ where $t = N_I\cdot M$ is the total number of patches (see \Cref{fig:tensor dictionary}).

To `learn' the dictionary representation is to minimize $\|\T{Y} - \TD * \T{H}\|_F^2$ where $\TD\in \Rbb_+^{p\times s\times q}$,  $\T{H}\in \Rbb_+^{s\times t\times q}$, and $s \ll t$.  Here,
$\T{H}$ contains the tensor coefficients for the tensor dictionary $\TD$. 

From \cite{Soltani2016}, the problem to solve is 
	\begin{align}\label{eq:admm tensor}
	&\min_{\TD, \T{H}, \TU, \TV} 
			\quad \frac{1}{2} \|\T{Y} - \TU * \TV \|_F^2 + \lambda \|\T{H} \|_{\text{sum}}
			+ I_{\Rbb_+^{s\times t\times q}}(\T{H}) + I_{\texttt{D}}(\TD)\\
	&\text{subject to}\quad \TD = \TU \quad \text{and} \quad \T{H} = \TV, \nonumber
	\end{align}
where $\TD, \TU \in \Rbb_+^{p\times s \times q}$ and $\T{H}, \TV \in \Rbb_+^{s\times t\times q}$.  In \Cref{eq:admm tensor}, $\lambda$ is a regularization parameter and the sum norm (see \Cref{defn:sum norm}) promotes sparsity of the coefficient tensor $\T{H}$.  
We denote the indicator function of a set $Z$ as $I_Z$.  Thus, $I_{\Rbb_+^{s\times t\times q}}(\T{H})$ ensures the coefficients $\T{H}$ are non-negative and $I_{\texttt{D}}$ ensures the dictionary $\TD$ belongs to the compact and convex set \cref{eqn:dictionary set}
	\begin{equation}\label{eqn:dictionary set}
	\texttt{D} \equiv \left\{\TD \in \Rbb_+^{p\times s\times q} \mid 
			\|\vec{\TD}_i \|_F \le \sqrt{pq},\, i = 1,\dots,s\right\}.
	\end{equation} 
As described in \cite{Soltani2016}, we impose the extra constraint that $\TD \in \texttt{D}$ (as opposed to $\TD\in \Rbb_+^{p\times s\times q}$) to avoid scaling ambiguity; that is, for any $\beta > 0$, $\|\T{Y} - (\beta\cdot \TD) * (\frac{1}{\beta}\T{H}) \|_F^2 = \|\T{Y} - \TD* \T{H} \|_F^2$.

We will not discuss the specifics of the tensor-based ADMM algorithm in this paper; we refer the reader to \cite{Soltani2016} for a full analysis.  When optimizing \cref{eq:admm tensor}, we project $\TD$ and $\T{H}$ into $\texttt{D}$ and $\Rbb_+^{s\times t\times q}$, respectively.  We choose to project $\TD$ into $\texttt{D}$ using the infinity norm, that is: 
	\begin{align}\label{eqn:projection}
	P_{\texttt{D}}(\TD)_{ij}^{(k)} = \min(\max(\TD_{ij}^{(k)},0),1),
	\end{align}
where $P_{\texttt{D}}$ is the projection operator, an option included in the publically available code \cite{Githubref}.

\subsection{Representation/Recovery Formulation}
Now suppose we have image $\MB \in \mathbb{R}_+^{N_r \times N_c}$ which we would like to represent in terms of a dictionary we learn by solving \cref{eq:admm tensor}.
Let $\TB$ be the $p\times M\times q$ be the patchified tensor representation, and $\TD$ the $p \times s \times q$ non-negative patch dictionary. 
To represent the non-negative image via our patch dictionary, we solve
	\begin{equation}\label{eqn:mrnsd formulation}
	\min_{\TC} \frac{1}{2} \|\TB - \TD * \TC \|_F^2 \quad 
		\text{subject to } \TC \in \Rbb_+^{s\times M\times q}.
	\end{equation}
In other words, if we can determine $\TC \ge 0$ such that $\TB \approx \TD * \TC$, then the image approximation is obtained by computing $\TD* \TC$ and mapping the resulting tensor back to a 2D image by inverting the patchification process.   But several issues warrant discussion before presentation of the algorithm to solve for $\TC$.  First, we need to give intuition as to why the tensor-based approach to the image model can provide significantly different results than the a matrix-based analogue, independent of the method produced to generate the dictionary.   Then, we need to consider choices of patch and dictionary sizes required to maximize the representation power and harness the computational efficiencies of the tensor-based approach.   These issues are covered in the next section.

\section{The Tensor Formulation: Advantages and Parameter Choices}  
\label{sec:superior}
We first explain the power behind the tensor-based approach.  Then, we discuss the choice of parameters such as 
dictionary size and patch size to maximize the potential of our new method. 

 \subsection{Tensor Superiority} 
 In this subsection, we will assume that a patch-dictionary has already been determined.   It does not matter for the moment
how that dictionary was derived:  our goal is to show the differences in the solution sets to the two problems of image approximation, one
based on a matrix-formulation, and one based on the tensor-formulation.   

Let $\TD \in \mathbb{R}_{+}^{p \times s \times q} $ denote the dictionary in tensor form, and define $\underbar{\MD} = \Unfold{\TD} \in \mathbb{R}_{+}^{pq \times s}$.   Likewise, let $\TB$ denote the patchified tensor image, and let $\underbar{\MB} = \Unfold{\TB} \in \mathbb{R}_{+}^{pq \times n_r n_c}$.   We have the following theorem:
\begin{theorem}\label{thm:superior}
Consider the set of solutions to within a tolerance $\epsilon$:
\[ \mathcal{X}_{mat} := \{ \M{C} \in \mathbb{R}_{+}^{s \times n_r n_c} | \| \underbar{\MB} - \underbar{\MD} \cdot \M{C} \|_F \le \epsilon \} \]
\[ \mathcal{X}_{ten} := \{ \TC \in \mathbb{R}_{+}^{s \times n_r n_c \times q} | \| \TB - \TD * \TC \|_F \le \epsilon \}. \]
Let $\mathcal{X}_{mat,e}$ denote the set of tensors of size $s \times n_r n_c \times q$ whose first frontal slice is from $\mathcal{X}_{mat}$ and the remaining $q-1$ frontal slices are zeros.   Then $\mathcal{X}_{mat,e} \subset \mathcal{X}_{ten}$.  That is, the set of solutions of the tensor problem effectively contains the set of solutions to the matrix problem. 
\end{theorem}  

\begin{proof} See \cite{LizThesis}. \end{proof}

This suggests that in solving the tensor problem, the solutions to the matrix problem are achievable, and we would 
be able to recover those if those are optimal in the tensor framework, as we demonstrate in \Cref{ex:one} below.  However, the
tensor case may provide better solutions by virtue of working in the tensor algebra, which we see in \Cref{ex:two}. 

\begin{example} \label{ex:one}
Let $\MB = \begin{bmatrix} 1 & 1 \\ 1 & 1 \end{bmatrix}$ be a single patch ($p=q=2$, $n_r=n_c=1$) which we interpret as the entire image, meaning $\underbar{\MB} = \mbox{\tt vec}(\MB)$ is $4 \times 1$.  Suppose
\[ \underbar{\MD} = \begin{bmatrix} 
	1 & 0 & 0 & 0 & \frac{1}{4} \\[0.25em] 
	0 & 1 & 0 & 0 & \frac{1}{2} \\[0.25em] 
	0 & 0 & 1 & 0 & \frac{3}{4} \\[0.25em] 
	0 & 0 & 0 & 1 & 1 \end{bmatrix}. \]
Set $\epsilon = 0$; that is, find exact solutions $\V{c} \in \Rbb_+^{5\times 1}$ such that $\| \underbar{\M{B}} - \underbar{\M{D}} \cdot \V{c} \| = 0$. 
It is easily checked that $\V{c}_a = [1,1,1,1,0]^T$ and $\V{c}_b = [3/4,1/2,1/4,0,1]^T$  with $\| \V{c}_a \|_{1} = 4$ and $\| \V{c}_b \|_{1} = 5/2$ are both exact solutions of the matrix optimization problem.  
It is also easy to see a non-negative solution cannot be obtained with fewer than four non-zero coefficients.   

We can exactly capture these matrix solutions in the tensor framework.  Let $\T{B} = {\tt twist}(\MB)$ be the patch stored as a $2\times 1\times 2$ lateral slice and let $\T{D} = \mbox{\tt fold}(\underbar{\MD})$ be the equivalent tensor dictionary of size $2\times 5\times 2$. 
We are trying to find exact solutions $\T{C}\in \Rbb_+^{5\times 1\times 2}$ such that $\| \T{B} - \T{D} * \T{C} \| = 0$. 
%
If we let $\TC_{:,:,1} = \V{c}_a$ and $\TC_{:,:,2} =  \V{0}$, then it is easily seen that $\TC$ is a solution of the tensor version of the problem -- this solution is effectively the matrix solution, in tensor form.  
However, the coefficient tensor
\[ \TC_{:,:,1} = [1/3,0,0,0,2/3]^T  \text{ and } \TC_{:,:,2} = [1/3,0,0,0,2/3]^T, \] 
is a solution to the tensor version of the problem with no matrix-based analogue, and we also observe $\| \TC \|_{\textnormal{sum}} = 2$.
Thus, the set of tensor solutions is bigger, and for the same number of non-zeros in the coefficients, we can get tensor solutions of smaller sum norm.  
\end{example}

\begin{example} \label{ex:two}
Here, we let $\MB = \begin{bmatrix} 1 & 3 \\ 2 & 4 \end{bmatrix}$, still assuming a single patch, and we assume 
\[ \underbar{\MD} = \begin{bmatrix} 1 & 0 & 0 & 1 & 1 \\ 0 & 0 & 1 & 1 & 1 \\ 0 & 0 & 1 & 0 & 1 \\ 0 & 1 & 0 & 0 & 1 \end{bmatrix}, \]
with $\TD = \mbox{fold}(\underbar{\MD})$.  It is easy to verify that ther is no non-negative $\V{c}$ that can exactly recover
$\mbox{\tt vec}(\MB)$ in the matrix case.   
If we set $\epsilon = 1/2$, then one element of $\mathcal{X}_{mat}$ is $\V{c} = [1/2,7/2,2,0,1/2]^T$ with $\| \V{c} \|_1 = 13/2$.   
However, with only four non-zeros entries in our tensor coefficients, we can resolve $\TB$ exactly: e.g.,   $\TC_{:,:,1} = [0,1,1,0,0]^T$ and  $\TC_{:,:,2} = [0,1,3,0,0]^T$ satisfies $\TB = \TD*\TC$, has four non-zeros, and has $\| \TC \|_{\text {sum}} =   6$.  

This demonstrates that we can get a richer and possibly more accurate set of solutions for the tensor representation of the problem than for the matrix version.   

\end{example}

 \subsection{Parameters}
 \label{subsec:localglobal}
In \cite{Soltani2016}, the authors use a value for $s$ that is consistent with a matrix-based patch dictionary learning algorithm, and illustrate on some CT image examples that when keeping $s$ fixed, the tensor patch dictionary allows for better approximation, but otherwise, the choices of $s$, $p$ and $q$ are not further discussed.  
Here, we explain why $s \ge p$ is necessary to get good representations.  We then discuss why $s \gg p$ is not advantageous from a storage perspective, and explain why $s=2p$ is sufficient from a qualitative point of view.   

\paragraph{\bf Dictionary Dimesion $s$}
In \cite{Soltani2016}, $s$ was chosen to be a small multiple of the product $pq$.   The reasoning for this was that in matrix patch dictionary learning, each patch is expressed as a vector and then approximated as a linear combination of the columns of the dictionary matrix.   Since the dictionary matrix would have $pq$ rows, then the choice of $s \ge pq$ would be required to try to ensure a spanning set.  However, taking $s$ this large for the tensor dictionary case is in fact not necessary for reasonably sized patches, as we now explain.   Further, large $s$ is not a good choice in terms of computational efficiency,  as we show later. 

From Definition (\ref{defn:tprodlat}), each of the $N$ image patches is approximated as 
\begin{equation} \label{eq:bi}  \MB_j \approx  \sum_{i=1}^s \MD_i \mbox{\tt circ}({\V{c}}_{ij}) , \qquad j = 1,\ldots,M, \end{equation}
where $\MB_j= \Squeeze{\vec \TB_j} $ and $\MD_i := \Squeeze{\vec \TD_i}$ are both in $\mathbb{R}_{+}^{p \times q}$ and ${\V{c}}_{ij} = \TC_{i,j,:}^T$.

Postmultiplication of $p \times q$ matrix $\MD_i$ by a $q \times q$ circulant generated by the tube $\V{c}_{ij}$ can be written 
\[ \MD_i \Circ{\V{c}_{ij}} = \V{c}_{ij}^{(1)} \MD_j  + \V{c}_{ij}^{(2)} \MD_j \MZ + \cdots + \V{c}_{ij}^{(q)} \MD \MZ^{q-1} , \]
where $\MZ$ denotes the $q \times q$ circulant downshift matrix (i.e. $\MZ^q = \MI$).   

Since each term in the sum (\ref{eq:bi}) admits such an expansion, after regrouping we obtain
\begin{equation} \label{eq:expand} \MB_j \approx \sum_{i=1}^s \V{c}_{ij}^{(1)} \MD_i + \sum_{j=1}^s \V{c}_{ij}^{(2)} \MD_i \MZ + \cdots + \sum_{j=1}^s \V{c}_{ij}^{(q)} \MD_i \MZ^{q-1} ,\end{equation}
meaning the $p \times q$ non-negative patch $\MB_j$ is described by a linear combination of $sq$, $p \times q$ non-negative matrices, although subsets of those matrices in the expansion are related via column shifts.   We know that a spanning set for all
$p \times q$ matrices would need to be of dimension $pq$.  We do not know if the matrices in the above expression are all independent so we do not know if $p=s$ is sufficient, but certainly we do need $s \ge p$.   We found in practice that it was sufficient to take $s$ a small multiple of $p$ as long as the patch sizes were not too large.  Typically $s = 2p$ was all that was needed in our experiments to get reasonable representations.   Though one might argue a larger value of $s$ may result in sparser coefficients, there is a trade-off with respect to the computational cost.

\paragraph{\bf Storage of Tensor Coefficients}
Storage of the original image requires storage of $N_r N_c$ pixel values.   
Storage of $\TD$ (assuming it is dense, which it may not be) requires $pqs$ numbers, while storage of $\TC$ requires
 $sq \frac{N_r}{p} \frac{N_c}{q} = s \frac{N_r}{p} N_c$ numbers.   Thus, if $s = 2p$, storage of $\TC$ {\it assuming that $\TC$ is dense} requires 
$2 N_r N_c $ numbers, twice the amount of storage of the image itself.   For the deblurring application, we will not be concerned with this additional storage -- the coefficients are a means to an end (namely, producing a high quality restoration).  For the compression application, however, our goal will be to produce a $\TC$ that is sparse, so that only non-zeros need to be stored.  

\paragraph{\bf Patch Sizes}
There clearly must be a lower bound on the patch size:  in the extreme with $p=q=1$, $s = 1$, the dictionary is only one non-zero constant and we cannot have compression because then $\TC$ is the image itself.    Choosing patch sizes too small undermines the power of the representation in (\ref{eq:expand}), and since the implementation of the algorithms utilizes FFT's of length $q$, there will be too much inefficiency if $q$ is very small (see further discussion in \Cref{subsec:patch size ratio}).  If $p$ is too large, since we have shown we need $s \ge p$, we would have a high storage cost for the dictionary.  After we discuss the algorithm, we will see that we are further constrained by the computational impact of the choice of patch sizes. 

\subsection{Global Interpretation: Image Resolution vs. Patch Size}

To gain intuition, we observe that by
placing all the image patches into position in the image, 
our tensor approximation is equivalent to the matrix representation 
\[  \MB \approx
\sum_{j=1}^s (\MI_{n_r} \otimes \MD_j) \begin{bmatrix}
\Circ{ \V{c}_{j,1} } & \Circ{\V{c}_{j,n_r+1}} & \cdots & \Circ{\V{c}_{j,n_r(n_c-1)+1}} \\ 
\Circ{ \V{c}_{j,2}}   &  \Circ{\V{c}_{j,n_r+2}} & \cdots & \Circ{\V{c}_{j,n_r(n_c-1)+2}} \\
\vdots & \vdots & \vdots & \vdots \\
\Circ{ \V{c}_{j,n_r}} & \cdots & \cdots & \Circ{\V{c}_{j,n_r n_c} } \end{bmatrix}, \]
where each circulant block in the block matrix is of size $q \times q$, and there are $n_r$ block rows and $n_c$ block columns.   
Thus, the image has a expansion in terms of a structured global dictionary $(\MI \otimes \MD_j), j=1,\ldots s$, although such an expansion is never computed explicitly.   From this we see that the same dictionary can be used to reconstruct the same image at different resolutions.  We illustrate this in the numerical results.  
\section{MRNSD for tensors}
\label{sec:mrnsd}

Since  
\[ \| \TB - \TD*\TC \|_F^2 = \| \mbox{\tt unfold}(\TB) - \mbox{\tt circ}(\TD)\mbox{\tt unfold}(\TC) \|_F^2 .\]   
let us (implicitly) define
\[ \V{v} = \mbox{\tt vec}(\mbox{\tt unfold}(\TB)), \qquad \V{c} = \mbox{\tt vec}(\mbox{\tt unfold}(\TC)), \qquad \mbox{ and } 
   \MD = \MI \otimes \mbox{\tt circ}(\TD). \] 
The MRNSD algorithm (\cite{Nagy2000,Kaufman1993}) was developed to solve $\min_{\V{c} \ge 0} \| \V{v} - \MD \V{c} \|_2 ,$ so it can clearly be applied to our formulation.  
    Of course it would be foolish to form $\MD$ explicitly.   In fact, we can use an equivalent and elegant formulation of each MRNSD step that uses all the tensor mechanics, and therefore only requires we have a
    routine that performs the $t$-product.   The algorithm is given in \Cref{alg:mrnsd}, and the details of the equivalence to this approach are given in  \Cref{sec:mrnsdderivation}.   

\begin{algorithm}[H]
\caption{MRNSD with $t$-product}
\label{alg:mrnsd}
\begin{algorithmic}[1]
\STATE{\textbf{Input:} image $\TB$, dictionary $\TD$, initial estimate $\TC_0$}
\STATE{Form gradient $\T{G}_0 = -\TD^T * (\TB - \TD * \TC_0)$}
\FOR{$k = 0,1,2,\dots$}
\STATE{$\T{S}_k = \TC_k \odot \T{G}_k$  
	\hfill \COMMENT{Form search direction (\Cref{sec:mrnsdderivation})}}
	
\STATE{$\theta_k = \texttt{trace}[(\T{S}_k^T * \T{G}_k)^{(1)}]/ \| \underbrace{\TD * \T{S}_k}_{\T{W}_k} \|_F^2$ 
		\hfill \COMMENT{Determine optimal step size (\Cref{sec:mrnsdderivation})}}
		
\STATE{$\alpha_k = \max\{\theta_k, \min_{\T{S}_{ij}^{(\ell)} > 0} (\TC_k)_{ij}^{(\ell)} / (\T{S}_k)_{ij}^{(\ell)}\}$
		\hfill \COMMENT{Ensure step size preserves non-negativity}}
\STATE{$\TC_{k+1} = \TC_k - \alpha_k \cdot \T{S}_k$ 
		\hfill \COMMENT{Update coefficients} \label{alg:update coefficients}}
\STATE{$\T{G}_{k+1} = \T{G}_k - \alpha_k \cdot \TD^T * \T{W}_k$ 
		\hfill \COMMENT{Update gradient}}
\ENDFOR
\end{algorithmic}
\end{algorithm}
%

\subsection{Implementation Details}
\label{subsec:patch size ratio}

Per iteration in \Cref{alg:mrnsd}, the dominant costs are the two products $\T{W}_k := \TD*\TS_k$ and $\TD^T *\T{W}_k$.  
Recall the $t$-product is computed by moving into the Fourier domain (i.e. computing $\widehat{\TD}, \widehat{\TS}_k$ and
and their facewise matrix-matrix products).   Some computations can be reused.  We note that $\widehat{\T{W}_k}$ need not be recomputed, since those entries are already known from computing $\T{W}_k$ in the step size computation.   Also, entries of $\widehat{\TD^T}$ are known from entries of $\T{D}$.  So we need only to assess the costs of computing 
$\widehat{\TD}, \widehat{\TS}_k$, and the costs of doing the $q$, matrix-matrix products $\widehat{\T{D}}^{(\ell)} \widehat{\T{S}}_k^{(\ell)}$ and $\widehat{\T{D}^T}^{(\ell)} \widehat{\T{W}}^{(\ell)}$.

 The computational cost for the FFTs is $O(s\cdot\log_2(q)\cdot(pq + \tfrac{N_r N_c}{p}))$, and 
the computational cost for the matvecs is $O( s N_r N_c )$.  

Note that the cost of the matrix multiplications is independent of the patch size if we assume serial implementation.  At the other extreme, for $q$ processors, each processor would compute a single matrix-matrix product at $s \tfrac{N_r N_c}{q}$ flops, so a larger value of $q$ is beneficial.  Either way, $q$ should not be too small or the constant in front of the cost to perform length-$q$ FFTs will not be suitably amortized. We already observed $s \ge p$.  If $s = k p$ for a small integer $k$,  the total flop count in serial is
$O( k p^2 \log_2(q) q + N_r N_ck (p+\log_2(q)) )$.  Note the cost grows more slowly for $q$, suggesting $p \le q$ may be desirable.  For sufficiently small fixed $k,p,q$, the cost grows as the number of unknowns in the image.


\subsection{Compression}
\label{subsec:compression}

When we reconstruct images via tensor MRNSD, we tend to generate coefficients $\TC$ which contain many small values.  This is because of the efficient encoding of information inherent in the $t$-product discussed previously.  

We introduce a sparsity regularization to our MRNSD minimization motivated by a proximal-operator framework~\cite{Boyd2013}.  
 We briefly outline the proximal gradient method; specific details can be found in~\cite{Boyd2013}. Traditionally, proximal algorithms are a class of convex optimization techniques solving problems of the following form:
	\begin{equation}\label{eqn:proximal framework}
	\min_{\V{c}} h(\V{c}) \equiv f(\V{c}) + g(\V{c}),
	\end{equation}
where $f$ is smooth and convex and $g$ is simple and convex.  For example, $f$ could be the $\ell_2$-norm (i.e., quadratic) and $g$ could be an $\ell_1$-regularization (i.e., piecewise-linear).  

We cannot use a conventional gradient-descent algorithm in \cref{eqn:proximal framework} because $g$ need not be differentiable.  Instead, we define the \emph{proximal operator} of $g$ as follows:
	\begin{equation}\label{eqn:prox operator}
	\text{prox}_g(\V{y}) = \argmin_{\V{c}} \left\{g(\V{c}) + \frac{1}{2} \| \V{c} - \V{y}\|_2^2\right\}.
	\end{equation}

The intuition behind \cref{eqn:prox operator} is to balance a point which minimizes a function $g$ that is close to another point $\V{y}$.  This interpretation gives rise to a two-step procedure to solve \cref{eqn:proximal framework}:

\medskip
	\begin{enumerate}
	\item Minimize $f$ using gradient descent: $\V{y} = \V{c}_k - \alpha_k \nabla f(\V{c}_k).$
		
	\item Find a nearby point which minimizes $g$: $\V{c}_{k+1} = \text{prox}_g(\V{y})$.
	\end{enumerate}
	
\medskip

We can apply this proximal operator framework to MRNSD with regularization.  For simplicity, we derive our method for matrix-vector products, with the understanding that we can translate this to tensor notation in our case, as we show at the end of this section.

Ideally, we use  $\ell_1$-regularization to promote sparsity in our original  problem:
	\begin{equation}\label{eqn:mrnsd with regularization1}
	\min_{\V{c} \ge 0} \frac{1}{2} \|\V{b} - \M{D}  \V{c} \|_F^2 + \lambda \| \V{c} \|_1.
	\end{equation}

Using MRNSD, we incorporate the non-negativity constraint into the optimization using the mapping $\V{c} = e^{\V{z}}$.  However, because $e^{\V{z}}$ is strictly positive, simply regularizing $e^{\V{z}}$ will not promote sparsity.  

We incorporate the constraint into our function using the following mapping:
	\begin{align*}
	\V{c} = e^{\V{z}} - \epsilon {\V{1}},
	\end{align*}
where $\V{1}$ denotes the vector of all ones. 
This means $\V{c}_i > -\epsilon$ and we will take $\epsilon \to 0$.  We now minimize the unconstrained problem:
\begin{align*}
	\min_{\V{z}} = \underbrace{\tfrac{1}{2} \|\V{b} - \M{D} (e^{\V{z}} - \epsilon) \|_F^2}_{f} + \underbrace{\lambda \|e^{\V{z}} - \epsilon \|_1}_{g}
	\end{align*}

We compute the gradient of $f$ and the proximal operator of $g$ as follows:
	\begin{align*}
	\nabla f 	&= e^{\V{z}} \odot [-\MD^T(\V{b} - \MD ( e^{\V{z}} - \epsilon))] \\
			&= (\V{c} + \epsilon) \odot [-\MD^T(\V{b} - \MD \V{c})],  \quad \epsilon \to 0. \\
			&= \V{c} \odot [-\MD^T(\V{b} - \MD \V{c})] 
	\end{align*}

This is the same gradient we had before.  Next we consider the proximal operator:
	\begin{align*}
	\texttt{prox}_g(\V{y}) 
		&= \text{arg}\min_{\V{c}} \left\{ \tfrac{1}{2}\|\V{c} - \V{y}\|_F^2 + \lambda \|\V{c} \|_1\right\}\\
		&= \text{arg}\min_{\V{z}} \left\{ \tfrac{1}{2}\|e^{\V{z}} - \epsilon - \V{y}\|_F^2 + 
				\lambda \|e^{\V{z}} - \epsilon \|_1\right\}
	\end{align*}
We solve this by computing the (sub)gradient and setting it equal to zero as follows:
	\begin{align*}
	e^{\V{z}}\odot (e^{\V{z}} - \epsilon- \V{y}) + \lambda e^{\V{z}}\odot \texttt{sign}(e^{\V{z}} - \epsilon) &= \V{0}.\\
	e^{\V{z}} - \epsilon- \V{y} + \lambda \cdot \texttt{sign}(e^{\V{z}} - \epsilon)  &=\V{0}.
	\end{align*}

We can map this back to $\V{c}$ as follows:
	\begin{align*}
	\V{c} - \V{y} + \lambda \cdot \texttt{sign}(\V{c}) &= 0.
	\end{align*}

The solution to the above equation is exactly the soft-thresholding operator.  Therefore, our MRNSD iteration with encorporated $\ell_1$-regularization is the following:
	\begin{align*}
	\V{c}_{k+1} = G_{\alpha_k \cdot \lambda}[\V{c}_k - \alpha_k \cdot \V{c}_k \odot (-\MD^T(\V{b} - \MD \V{c}_k)) ],
	\end{align*}
where $G_\mu$ is the soft-thresholding operator:
	\begin{align*}
	G_{\mu}[c] &= \left\{ \begin{array}{ll} 
		c - \mu, & c > \mu \\
		0, & |c| < \mu \\
		c + \mu, & c < -\mu.
		\end{array}\right.
	\end{align*}

Using the same observations as in the start of this section that allowed us to move from the matrix formulation to the tensor formulation, 
we arrive at the sparsity-constrained tensor-MRNSD formulation by changing \Cref{alg:mrnsd}, Line \ref{alg:update coefficients} to the following:
	\begin{align}\label{eq:update coefficients with sparsity}
	 \TC_{k+1} = G_{\alpha_k \cdot \lambda}[\TC_k - \alpha_k \cdot \T{S}_k].
	\end{align}

We conclude this section by noting that we are not the first to consider augmentation of MRNSD iterates in order to encourage sparsity.   In \cite{Cickbook}, the authors suggest applying a sparsity-type constraint to an MRNSD step.  But the text was without mathematical justification, and we found in our examples that incorporation of their suggestion did little to promote sparsity.


\section{Deblurring}
\label{sec:deblur}

 We briefly review the standard model for image blurring/deblurring to set the stage for our tensor-based deblurring approach.
For more background see \cite{OlearyEtAl}.   

The basic blurring model given assuming known image $\V{x}_{true} = \mbox{\tt vec}(\MX_{true}) \in \mathbb{R}^{N_r \times N_c}_{+}$, 
is $$\MA \V{x}_{true} + \V{n} = \V{b},$$ 
where $\MA \in \mathbb{R}^{N_r M_r \times N_c M_c}$ is a blurring operator whose singular values decay rapidly to 0, $\V{n}$ is the unknown white noise vector and $\V{b}$ is the blurred noisy image in vector form; that is, 
$\V{b} = \mbox{\tt vec}(\MB)$ where $\MB$ is $M_r \times M_c$.   Since $\MA$ and $\V{b}$ are known but the noise is not, one might be tempted to ignore the noise, and compute the minimum-norm, least squares solution to $\MA \V{x} = \V{b}$.  However, the ill-conditioning of the operator renders the least squares solution worthless, since small singular values magnify the noise present in the data.   

Algorithms for computing estimates $\V{x} \approx \V{x}_{true}$ in the presence of noise are called regularization methods.   Iterative solvers, such as MRNSD, can be used as regularization methods.  Consider applying MRNSD to 
   \[ \min_{\V{x} \ge 0} \| \V{b} - \MA \V{x}\|_2. \]
It will produce sequences of iterates $\V{x}_k$.  Those iterates tend to exhibit semi-convergent behavior in that they will approximate the noise-free solution $\V{x}_{true}$ with increasing $k$, up to a point.  
After a particular iteration, the method starts to fit the noise in $\V{b}$ to solve the optimization problem, and the solution begins to resemble the noise-contaminated solution.   If the stopping parameter is picked before the contamination happens, the method is considered to be a regularization method.

In our approach, we require non-negativity of the image estimate in addition to the fact that the image be comprised from a learned, non-negative patch dictionary.   In other words, we want our image estimate (expressed as a tensor, $\TX$) to be given by $\TX \approx \TD * \TC_k $ for $\TC_k \ge 0$.    
MRNSD can be used to treat this problem.  But first we need to show that it is possible to express the term on the right in the norm via matrix-vector products (but yet still employ the tensor format for computational efficiency during actual implementation).   Consider the relationship between the two formats of the same image: $\V{x}$ and $\TX$.
We see that 
\begin{equation} \label{eq:Pdef}  \V{x} = \MP \mbox{\tt vec} (\Unfold{ \TX} ) \approx \MP \mbox{\tt vec}(\Unfold{ \TD*\TC }),  \end{equation}
for a permutation matrix $\MP$.  
Now $\Unfold{\TD * \TC} = \underbar{\MD} \, \Unfold{\TC}$, where $\underbar{\MD} = \Circ{\TD}$.  
So $\mbox{\tt vec}(\Unfold{\TD*\TC}) = (\MI \otimes \underbar{\MD}) \mbox{vec}(\Unfold{\TC})$.  
Thus, we can apply MRNSD to solve
\[ \min_{\TC \ge 0} \| \V{b} - \left(\MA \MP (\MI \otimes \underbar{\MD}) \right) \mbox{\tt vec}(\Unfold{\TC}) \|_2 \]
though in practice, we construct neither $\MP$ nor $\MI \otimes \underbar{\MD}$ explicitly, since all the necessary computations can be done with permutation indicies and $t$-products with $\TD$ and $\TD^T$.    The computational cost of 
one iteration is dominated by matrix-vector products with $\MA, \MA^T$ and products with $\TD, \TD^T$, which as we saw 
previously, for sufficiently small values of $p, q$ is a small multiple of the number of unknowns in the image (and can be efficiently parallelized).

\section{Numerical Experiments}
\label{sec:num}

We illustrate the power of the tensor dictionaries to represent images, both qualitatively and quantitatively.  
In all examples, we represent square images and the dimensions of the image and patches are powers of two.

\subsection{Power of Tensor Representations}
As discussed in \Cref{sec:superior} and \Cref{thm:superior}, the tensor representations can exactly capture the matrix representations and there are a greater number of possible tensor representations.  
To illustrate the advantages of using a tensor representation, we compare the representations with either a learned tensor dictionary $\T{D} \in \Rbb_+^{16\times 32\times 16}$ against representations from a learned matrix dictionary $\M{D}\in \Rbb_+^{256\times 512}$.  
In both cases, we use patches of size $16\times 16$, either stored as lateral slices of $\T{D}$ or as columns of $\M{D}$. 
The number of dictionary elements in each case is twice the size of the first dimension (i.e., the dictionaries are equally over-complete).   
Both dictionaries were formed solving the  ADMM  formulation from images of faces in the CalTech101 database \cite{Caltech101}.

\begin{figure}[ht]
	\centering
	\def\n{4} 
	\subfigure[Original: $\M{B} \in \Rbb_+^{512\times 512}$. \label{subfig:dog original}]{\includegraphics[height=\n cm]{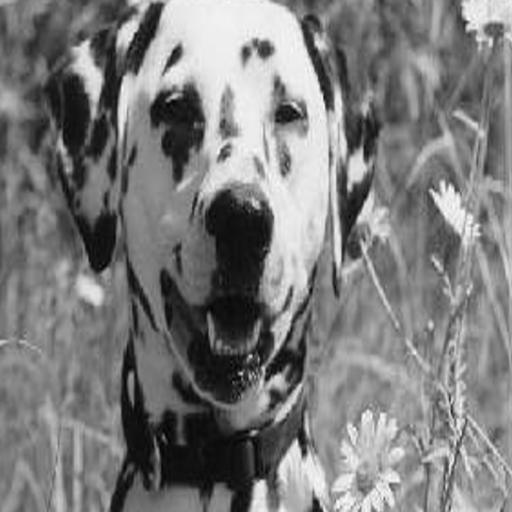}}
	\subfigure[Ten: $\frac{\| \T{B} - \T{D} * \T{C} \|}{\| \T{B}\|} \approx 0.03$ 
		\label{subfig:dog tensor}]
		{\includegraphics[height=\n cm]{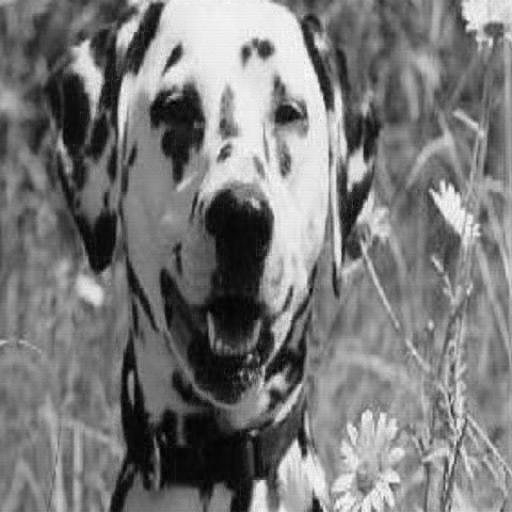}}
	\subfigure[Mat: $\frac{\| \underline{\M{B}} - \M{D} * \M{C} \|}{\| \underline{\M{B}}\|} \approx 0.10$ 
			\label{subfig:dog matrix}]
			{\includegraphics[height=\n cm]{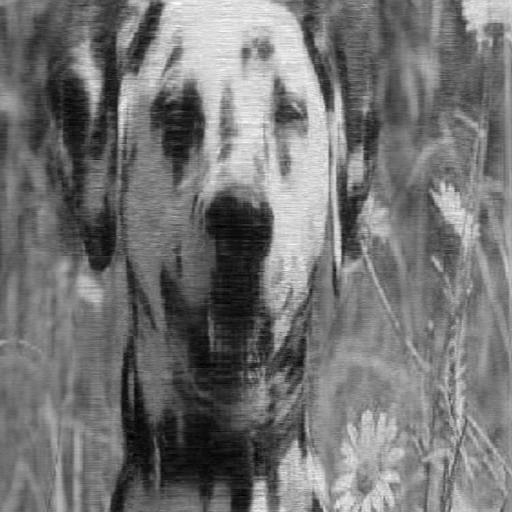}}
	
	\caption{Comparison of tensor representation \ref{subfig:dog tensor} with learned $\T{D}\in \Rbb_+^{16\times 32\times 16}$ vs. learned matrix representation \ref{subfig:dog matrix} with $\M{D} \in \Rbb_+^{256\times 512}$.}
	\label{fig:matrix vs tensor}
	\end{figure}

We form our representations in \Cref{fig:matrix vs tensor} using $200$ MRNSD iterations (\Cref{alg:mrnsd}) and we start with a random, normalized initial guess. 
	
In \Cref{fig:matrix vs tensor}, we see the tensor representation in \ref{subfig:dog tensor} is better than the matrix representation in \ref{subfig:dog matrix}, both numerically and qualitatively.  
Thus, not only are there more possible the tensor representations (see \Cref{thm:superior}), the representation we form is better.  
This is somewhat surprising as the number of coefficients (i.e., the representation ability) for both the tensor and matrix cases is the the same.
More specifically, the sizes are the following: the tensor coefficients $\T{C} \in \Rbb_+^{32\times 1024\times 16}$ and the matrix coefficients $\M{C} \in \Rbb^{512 \times 1024}$ where $1024$ is the number of patches in our original image $\M{B}$.  
A potential reason for this improved representation is that the patches stored in the tensor dictionary $\T{D}$ maintain some spatial relationships typical in natural images (e.g., smooth curves) whereas the patches stored in the matrix dictionary $\M{D}$ are more binary (e.g., sharp edges).  


\subsection{Fixed Dictionary, Changing Resolution}
\begin{figure}[htb]
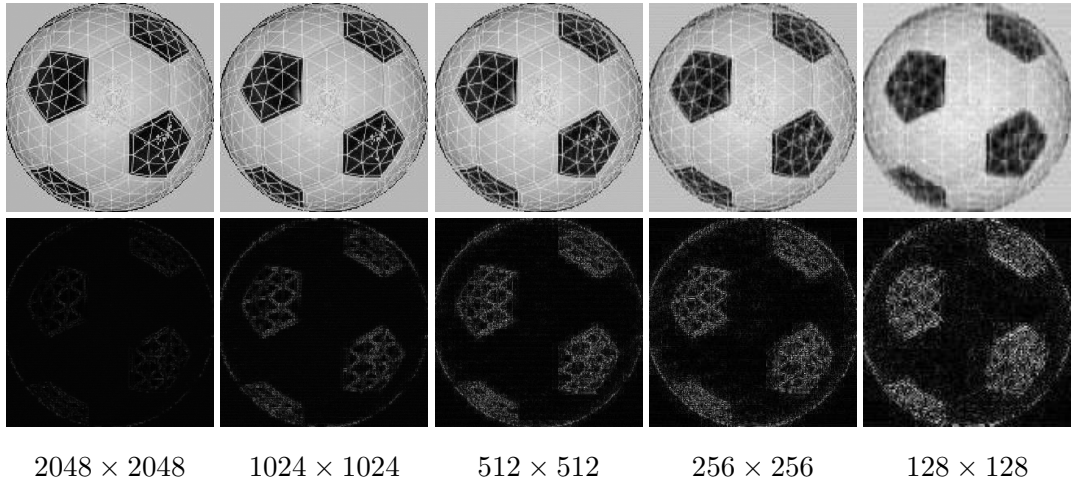

\centering
\def\n{2.75} 

\begin{tikzpicture}
\def\N{2.85}
\foreach \i/\size in {0/2048, 1/1024, 2/512, 3/256, 4/128}{
	\node at (\i*\N, 0) {\includegraphics[height=\n cm]{soccer_recon_\size.jpg}};
	\node at (\i*\N, -\N) {\includegraphics[height=\n cm]{soccer_diff_\size.jpg}};
	\node at (\i*\N, -1.5*\N-0.5) {$\size \times \size$};
}
\end{tikzpicture}

\caption{Effects of representations with $\T{D}\in \Rbb_+^{16\times 32\times 16}$ as the image resolution changes.  
The dictionary was formed from $128 \times 128$ images. 
The top row shows the representations $\T{D}* \T{C}$ and the bottom row shows the absolute difference $|\T{B} - \T{D} * \T{C}|$.  
The resolution decreases from left to right.}
\label{fig:scale} 
\end{figure}

As noted previously, independent of how the dictionary was learned, we can employ that dictionary (assuming appropriate dimensions) on multiple resolutions of the same image, as illustrated in \Cref{fig:scale}.    
Importantly, the fact that  the data was trained on images of a different resolution (in this case, the training data were all $128 \times 128$ images) is insignificant.

In \Cref{fig:scale}, we notice that as the image resolution decreases, the quality of our representations decreases as well (this is borne out by our reconstruction relative error). 
This is because the size of our patch relative to the image increases; i.e., each patch is representing a larger portion of the image, and hence is less likely to match exactly. 
From another perspective, when we represent an image with a higher resolution, the dictionary patches act more like individual pixels in the image and hence provide a more accurate representation.

\subsection{Color}  
\begin{figure}[htb]
\centering
\def\n{4} 
\subfigure[Original $512\times 512$, $\M{B}$.]{\includegraphics[height=\n cm]{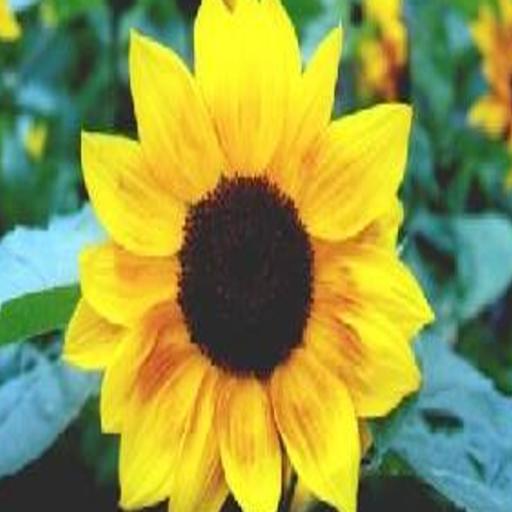}}
\subfigure[Representation,  $\TD * \TC$.]{\includegraphics[height=\n cm]{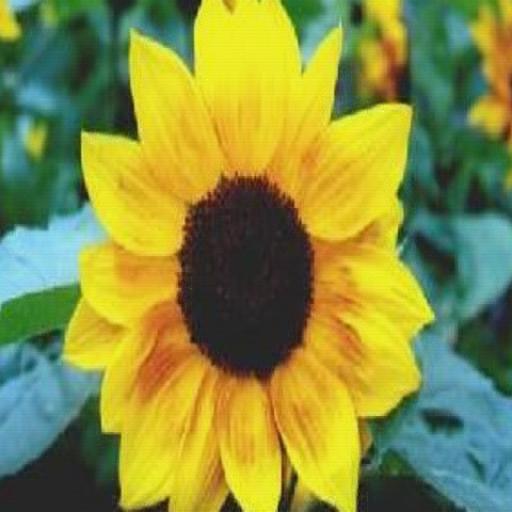}}
\subfigure[Difference, $|\TB - \TD * \TC|$.]{\includegraphics[height=\n cm]{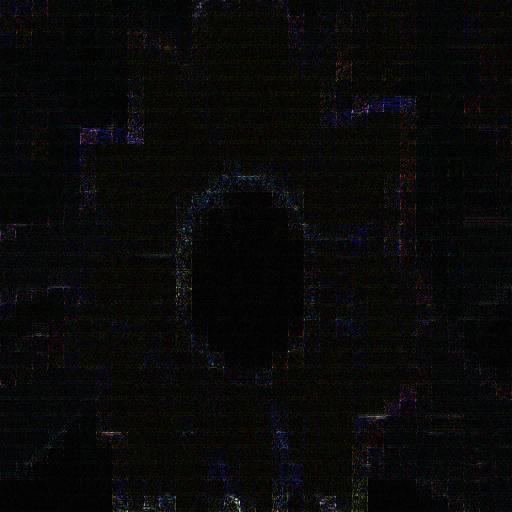}}
\caption{Representing color images using $\T{D}\in \Rbb_+^{16\times 32\times 16}$.  The relative error of our representation is $\| \T{B} - \T{D} * \T{C} \| / \| \T{B} \| \approx 0.02$.}
\label{fig:color}
\end{figure}

We can also represent color images using the same dictionary generated from grayscale images.  
Suppose we have an RGB image of size $N_r\times N_c \times 3$ where the third dimension is the number of color channels.  
To patchify an RGB image, we treat each channel as a separate grayscale image from which we form patches and store as lateral slices of a tensor.  
This means we have three tensors of size $p\times M \times q$.
We then concatenate the lateral slices of the patchified tensors for each color channel to obtain our RGB patchified tensor of size $p\times 3M\times q$.  
In \Cref{fig:color}, we depict a representation of a color image using the same tensor dictionary $\T{D}\in \Rbb_+^{16\times 32\times 16}$.

\subsection{Compression}

In the examples, the $\M{B}$ is $N \times N$.   If we want to talk about the compression of a single image via the approximation $\TB \approx \TD*\TC$, we need to compute the compression ratio
\[ \frac{ \mbox{nnz}(\TD) + \mbox{nnz}(\TC)}{ N^2 } .\]
However, if we are storing compressed representations of multiple images where they have all been compressed using the same 
dictionary, the cost of storing the dictionary becomes amortized over the multiple test images, so we approximate compression via $\mbox{nnz}(\TC) / N^2$.  

For a fixed patch size, we know we want $s \ge p$.   We can make $s$ larger (maybe a bit larger than $2p$), and increase sparsity to a point, but too big $s$ means too much non-uniqueness and the optimization problem gets trickier.   
We can change patch size.  Increasing patch size for a fixed resolution beyond a certain point is not a good idea -- we lose representability.  But for larger images, we may well want to increase the patch size if we think our representation may be more sparse and we don't lose much representability.  If we do that, $s$ must increase as a small multiple of $p$ and the cost of producing $\TC$ increases.

\begin{figure}[htb]
\centering 
\includegraphics[width=\textwidth]{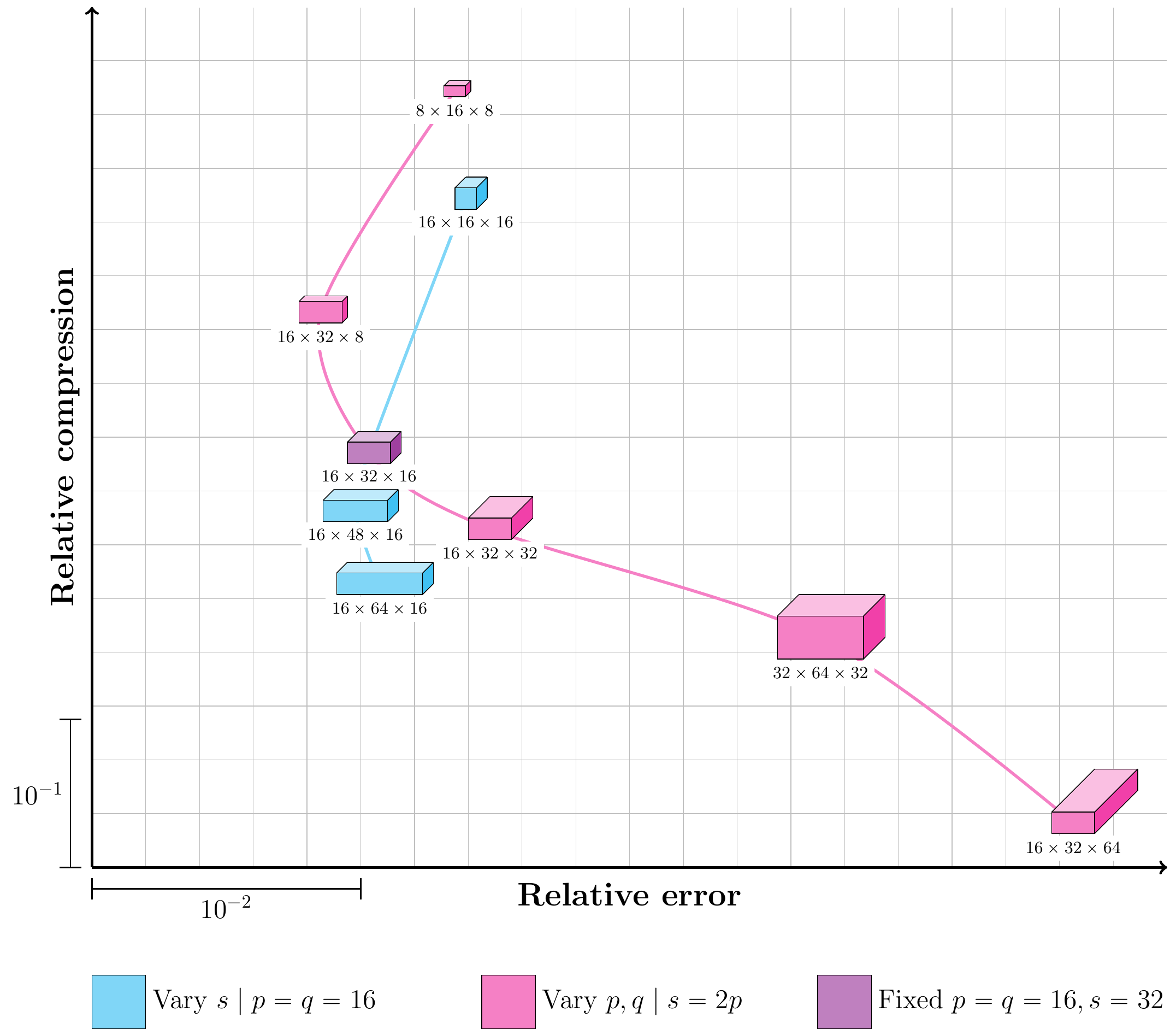}

\caption{Comparison of approximate compression for various dictionary sizes using sparsity-promoting MNRSD (\Cref{alg:mrnsd} with  \eqref{eq:update coefficients with sparsity}) and a sparsity parameter of $\lambda=10^{-10}$. 
The cyan line represents dictionaries of the same patch size, but varying the number of dictionary elements (i.e., width).  
The magenta line represents dictionary of various patch size, but the same level of of over-completeness (i.e., twice as many lateral slices as the first patch dimension $p$).}
\label{fig:compression with dictionary size}
\end{figure}

To examine the effects of patch size on compressibility, we compare the relative error to the approximate compression $\mbox{nnz}(\TC) / N^2$ where $N = 512$ in \Cref{fig:compression with dictionary size}.  We use $200$ MRNSD iterations (\Cref{alg:mrnsd}) with the soft-thresholding step \eqref{eq:update coefficients with sparsity}.

There are a few key trends to notice in \Cref{fig:compression with dictionary size}.  The first is that the more over-complete a dictionary is, the more compressed the representation without significant loss of accuracy (the cyan dictionaries).  
This behavior occurs because with a wider selection of dictionary patches to select, we likely need to select fewer patches to represent an image well. 
However, if we include the cost of storing these wider dictionaries, the compression ratio greatly increases due to the width of the dictionary.    

The second trend is that the larger the patch size, the more compressed the representation, however with a significant loss of accuracy (the magenta dictionaries).   
This behavior occurs because larger patches are able to capture larger sections of an image, hence fewer patches are required form the representation.  
However, the larger patches are less likely to reproduce the original image exactly, and hence decreases the representation quality.  
Interestingly, if we include the cost of storing the dictionaries with large patches, it does not significantly impact the overall storage cost -- the width of the dictionary relative to the first patch size $p$ is a substantially more significant factor.



\subsection{Deblurring Results}

We used Matlab and features in the RestoreTools Matlab toolbox \cite{restoretools} as indicated.  

\subsubsection{Example 1}

\begin{figure}[H]
\centering
\subfigure[Original.]{\label{fig:aa1} \includegraphics[scale=.2]{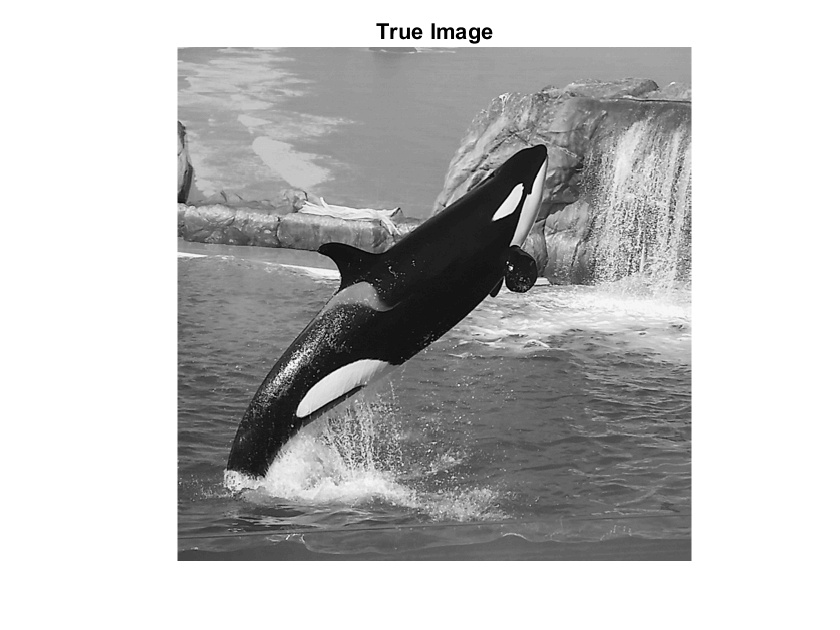}}
\subfigure[Blurred, noisy.]{\label{fig:bb1} \includegraphics[scale=.2]{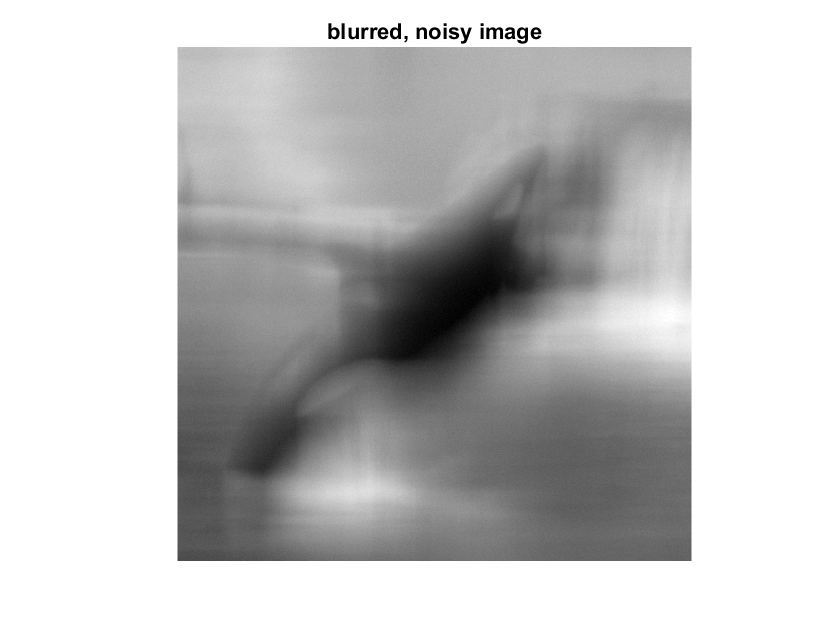}}
\subfigure[Ten. rcn $p,q=16$, opt]{\label{fig:cc1} \includegraphics[scale=.2]{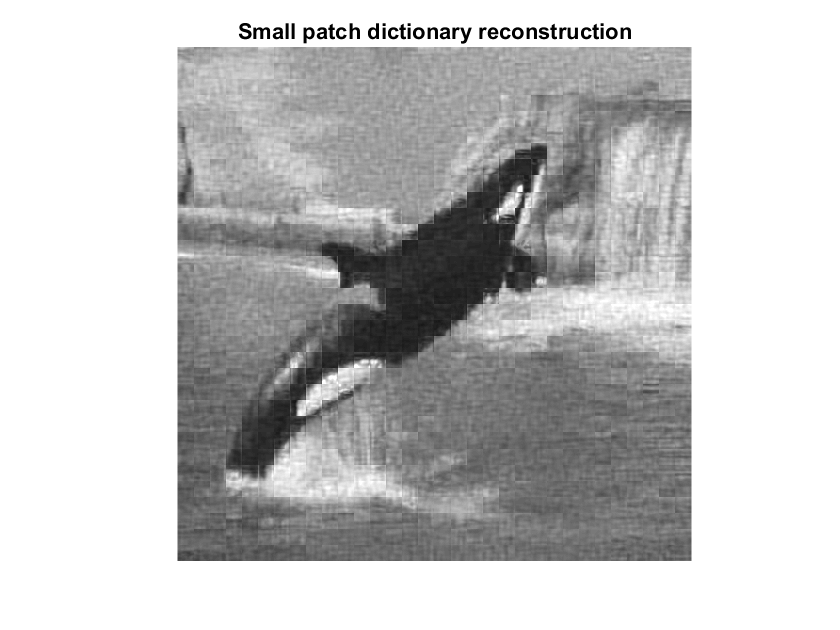}}
\subfigure[Ten. rcn. $p,q=32$]{\label{fig:dd1} \includegraphics[scale=.2] {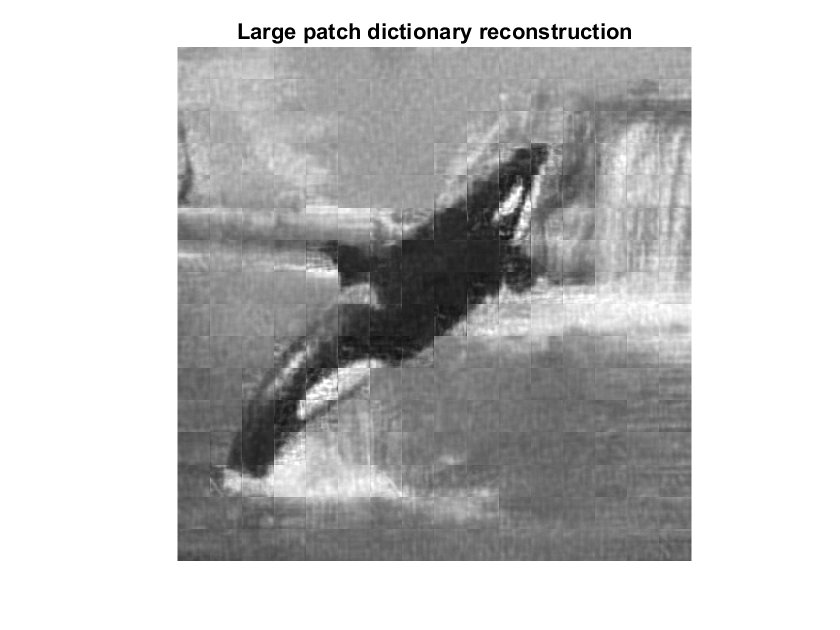}}
\subfigure[Combined tensor]{\label{fig:ee1} \includegraphics[scale=.2]{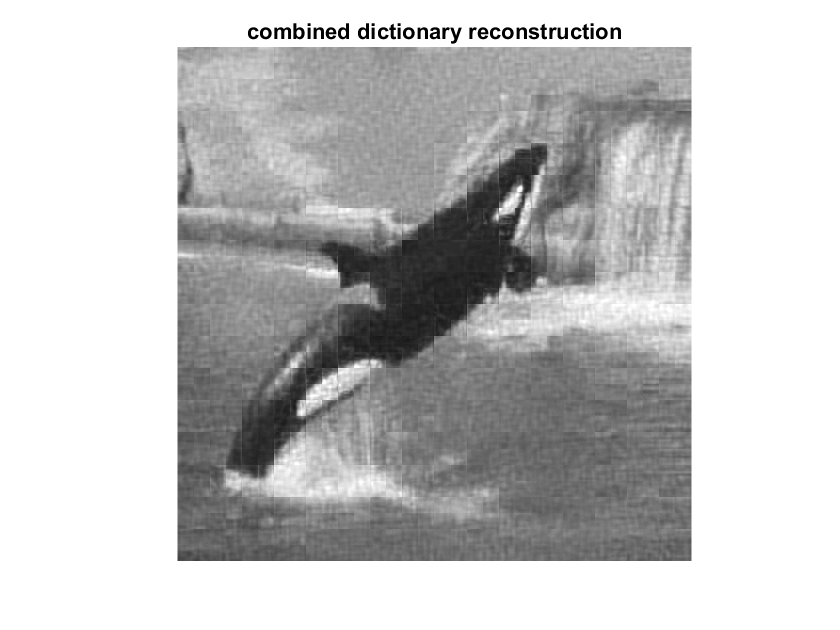}}
\subfigure[Matrix Recon, opt]{\label{fig:ff1} \includegraphics[scale=.2]{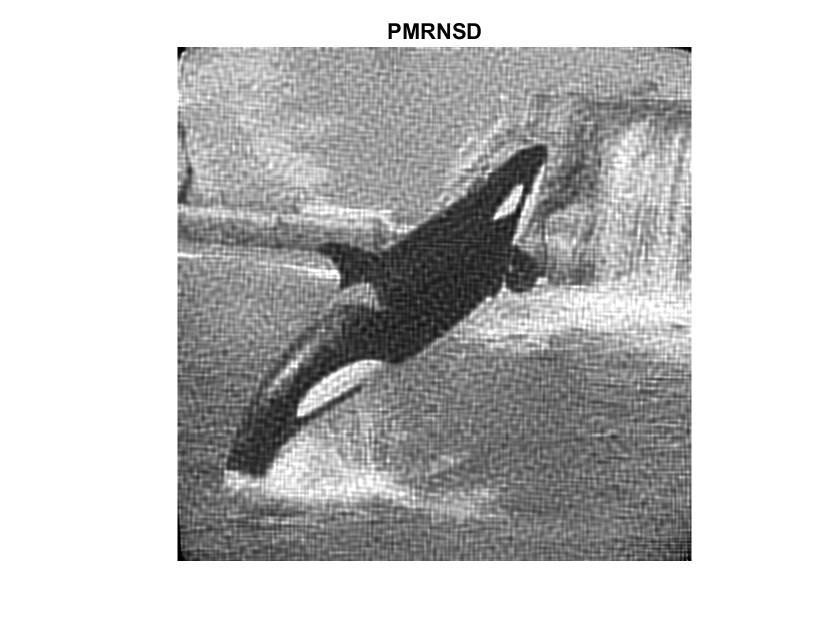}}
\caption{Example 1:  Orca original, blurred and noisy images, and various reconstruction results. }
\label{fig:orcarecon}
\end{figure}

Our true image was the $512 \times 512$ images of the orca in \Cref{fig:aa1}.   We used the grain blur in 
Restoretools set of example files\footnote{The grain blur point-spread-function is for a 256 x 256 image, so we padded the blur by zeros to get a PSF suitable for a 512 x 512 image.} to create a blurring operator $\MA$ corresponding to reflexive boundary conditions.    We computed $\MA \V{x}_{true}$, and added Gaussian noise at a noise level of 1 percent to the image. The blurred, noisy image in the figure.  

Convergence to regularized solutions is known to be slow with MRNSD \cite{Nagy2000}, so preconditioning is often used. Thus, in both the non-dictionary and dictionary reconstructions, we used the built-in preconditioner option and used MRNSD on the preconditioned problems
  \[ \min_{\V{x} \ge 0} \| \M{M} \V{b} - \M{M} \MA \V{x} \|_F  \mbox{ or }
\min_{\TC \ge 0} \| \M{M} \V{b} - \M{M} \left(\MA \MP (\MI \otimes \underbar{\MD}) \right) \mbox{\tt vec}(\Unfold{\TC}) \|_F \]
where $\M{M}$ denotes the preconditioner determined from the PSF and $\V{b}$, using the default settings. The matrix $\MP$ is a permutation matrix (see (\ref{eq:Pdef}).

The algorithm needs a non-zero starting guess.  In the matrix case, we used a vector of all ones as the initial guess for $\V{x}$.   In the tensor case, to make an equivalent comparison, we first formed a patchified version of an image of all ones.  We then multiplied that by the 
tensor-pseudoinverse of $\T{D}$ (see \cite{KilmerEtAl2013} for details) and used this for the starting guess for $\T{C}$.

We wanted to compare the quality of MRNSD with and without dictionaries.   We do not discuss choosing optimal truncation parameters, though we note that the semi-convergence behavior is very much damped when using the dictionaries.  We used two dictionaries derived from 
different data sets at different patch sizes.  The first dictionary was obtained from the CalTech face database.  We took $p=q=16$ and $s=32$.  The second dictionary was obtained from a collection of 60 elephant photos \cite{Caltech101}.  Here, we took $p=q=32$ and $s=64$.   

In the figure we compare the `optimal' (i.e. solution at the iterate that gave smallest relative error against ground truth) solution with preconditioned MRNSD with no dictionary approach against other reconstructions.  In \Cref{fig:cc1}, we give the optimal reconstruction for the smaller dictionary.  In \Cref{fig:dd1}, we give the solution after 2000 iterations for the larger dictionary (the error is still decreasing at this point, so it may not be an optimal stopping point).  In \Cref{fig:ee1}, we averaged the solutions\footnote{In fact, any convex combination of the reconstructions would have been an option.} to acknowledge the fact that this image has both fine scale features and components that are nearly uniform, so we expect that different resolution patches would be sensitive to this fact.  We address the issue of multiresolution reconstructions in the Conclusions.

All dictionary based solutions gave reconstructions with smaller relative error and smaller structured similarity as shown in the table.  It is worth noting that the dictionary-based reconstructions took longer to converge:  while preconditioned MRNSD in the matrix-only case took 63 iterations to reach the optimal solution, it took the small dictionary 1,211 iterations, and as mentioned, we let the large dictionary case run 2000 iterations.   On the other hand, this is not an entirely fair comparison, either, since the preconditioner was constructed relative to $\MA$, whereas in the matrix-formulation of our tensor approach, we see the structure of the matrix-operator is quite a bit different.      

\begin{table}[H]
\centering
\begin{tabular}{|c|c|c|c|c|}   \hline 
 & Small Dictionary & Large Dictionary & Combined & PMNRSD \\ \hline
Rel Err & 0.119   & 0.123  &  0.116 &  0.144\\ \hline
SSIM & 0.518  &  0.522  &  0.541 & 0.376 \\ \hline
\end{tabular}
\caption{Relative error and SSIM results for Example 1.   The combined-dictionary image had slightly better relative error and SSIM results than any other.  The matrix-based, non-dictionary reconstruction has notably worse relative error and SSIM to all the other dictionary-based reconstructions.}
\end{table}

\subsubsection{Examples 2 and 3:  Underdetermined Problems}

In the first illustration, our true image was $256 \times 256$.   We wanted to simulate a situation in which the boundary conditions of the blur were taken to be unknown.   We took a symmetric Gaussian blur of discrete bandwidth 8 and $\sigma = 3$, and applied it to the true image, trimmed blurred the image by 8 pixels on all sides, reshaped, and added 1 percent random Gaussian noise to the data\footnote{To implement this process in {\sc Matlab}: Let $\V{v} = \mbox{exp}(-\frac{1}{\sqrt{2\sigma}}[0\!\!:\!\!7]\mbox{.}^2)$, and $\MA_1 = {\tt toeplitz}(\V{v})$.   Define $\M{T} = \MA_1 * \MX_{true} * \MA_1$, and $\V{b}_{true} = \M{T}(8\!:\!247,8\!:\!247)$, $\V{b}_{true} = \V{b}_{true}(:); $ 
and $\V{b} = \V{b}_{true} + c \cdot \mbox{\tt randn}( {\tt length}(\V{b}_{true},1  ))$ where $c$ is such that the noise level is 0.01.} .  This meant the
data vector was only length $240^2$ while the true image was $256^2$, indicating there are fewer equations than unknowns.

Since the problem is underdetermined, it may be desirable to add regularization to enforce smooth transitions between patches.  For an $N \times N$ image and $p \times p$ patches,  we consider  
\[ \min_{\TC \ge 0} \left\| \bea{c} \V{b} \\ 0 \eea - \left( \bea{c} \MA \\ \lambda \ML \eea
\MP (\MI \otimes \underbar{\MD}) \right) \mbox{\tt vec}(\Unfold{\TC}) \right\|_2,  \qquad
 \ML = \bea{c} \MI \otimes \MQ \\  \MQ \otimes \MI \eea,   \qquad
 \]
where $\MQ$ could either be an $(N-1) \times N$ first order discrete derivative operator, or, in order to minimize computation, an $(\frac{N}{p} -1) \times N$ matrix approximating discrete derivatives only across patch jumps.   We expect, for a suitable value of $\lambda$, some smoothing across patch boundaries.  As our results below show, there is some modest gain that can be had when including extra regularization. However our method is relatively insensitive to choice of $\lambda$, whereas 
MRNSD without regularization is not.

\begin{figure}
\centering
\subfigure[Original.]{\label{fig:a1} \includegraphics[height=1.5in,width=1.6in]{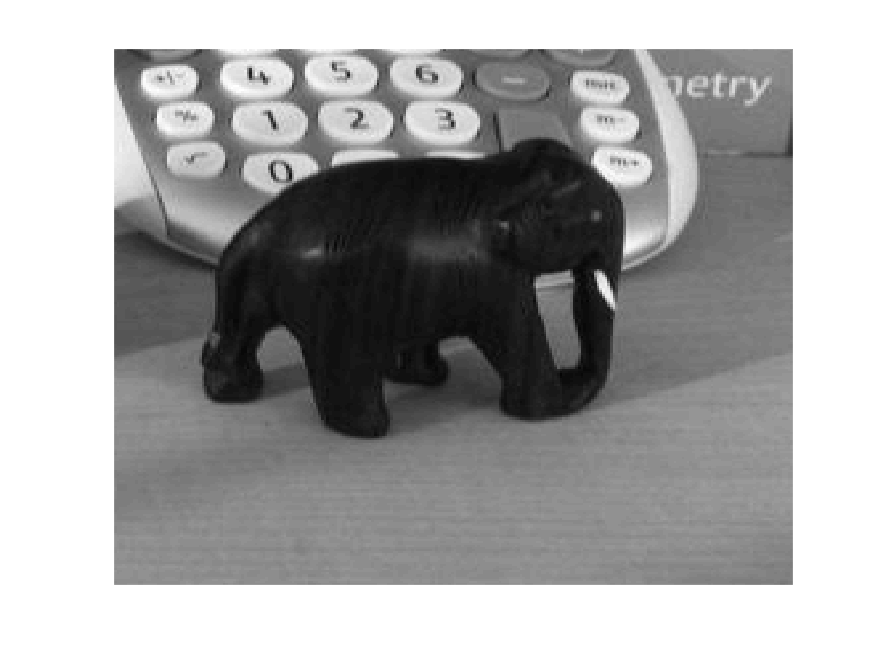}}
\subfigure[Blurred, noisy.]{\label{fig:b1} \includegraphics[height=1.5in,width=1.6in]{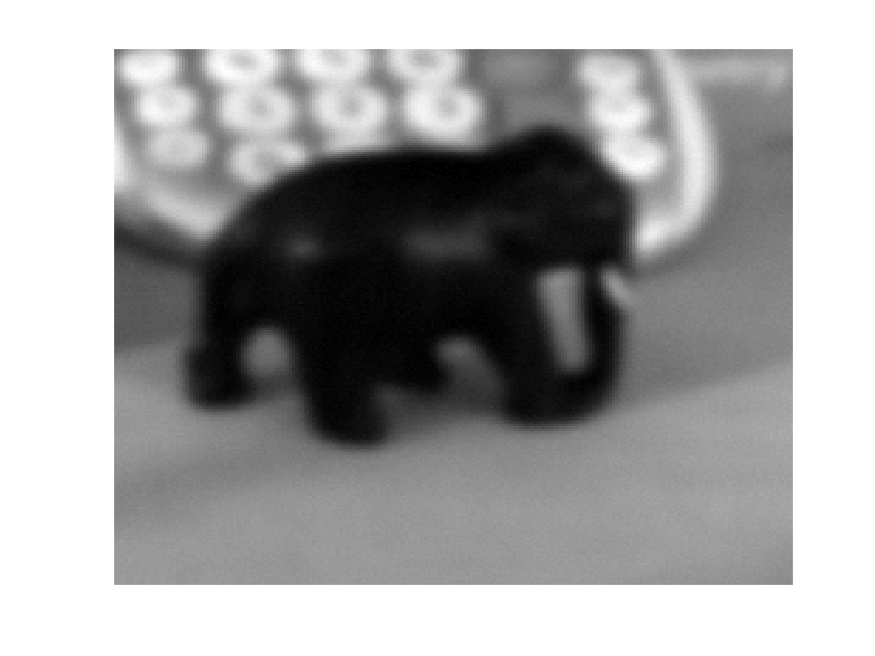}}
\subfigure[Ten rcn $\lambda=10$]{\label{fig:c1} \includegraphics[height=1.5in,width=1.6in]{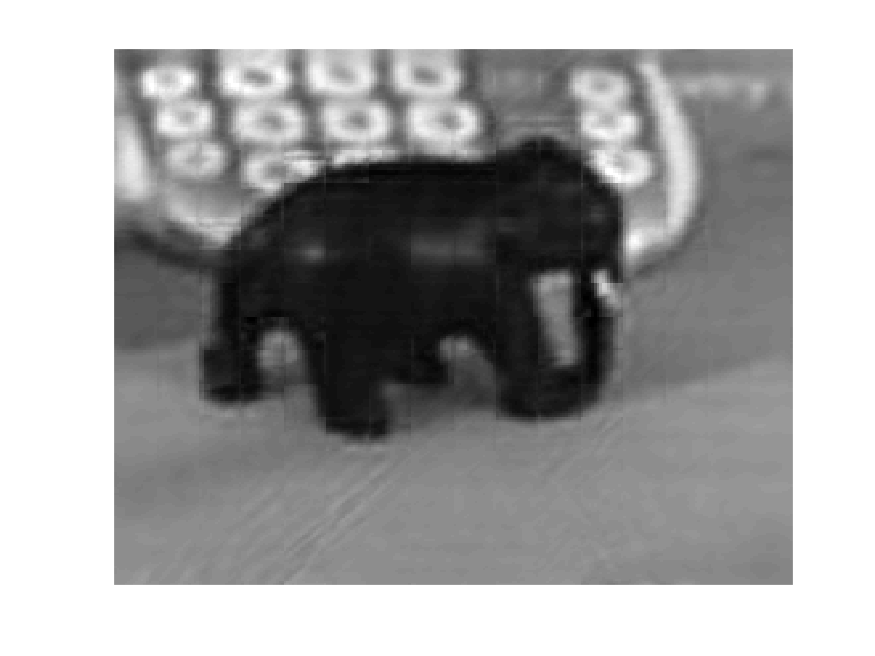}}
\subfigure[Ten rcn $\lambda = 0$]{\label{fig:d1} \includegraphics[height=1.5in,width=1.6in] {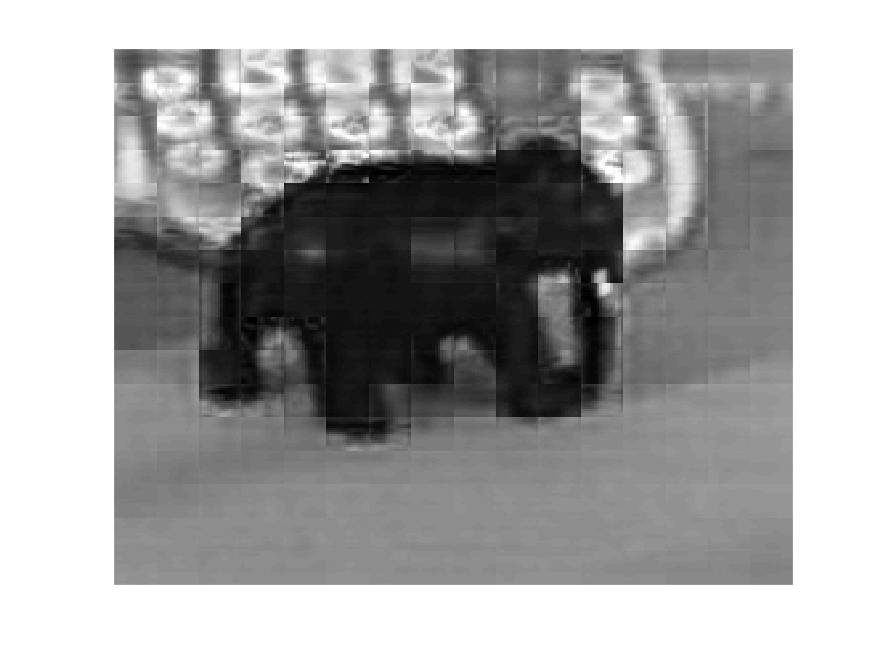}}
\subfigure[Matrix rcn, $\lambda=10$]{\label{fig:e1} \includegraphics[height=1.5in,width=1.6in]{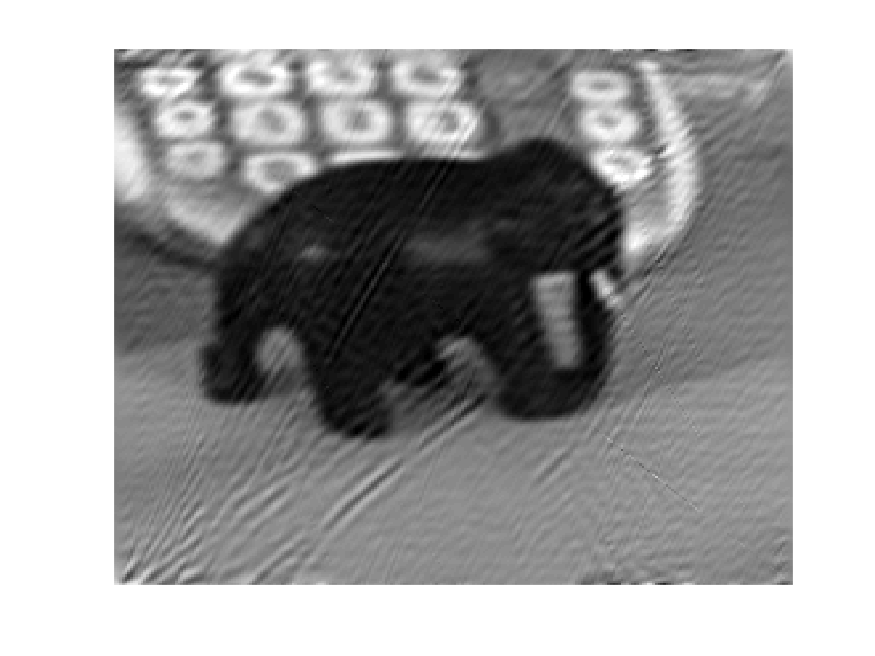}}
\subfigure[Matrix rcn, $\lambda=0$]{\label{fig:f1} \includegraphics[height=1.5in,width=1.6in]{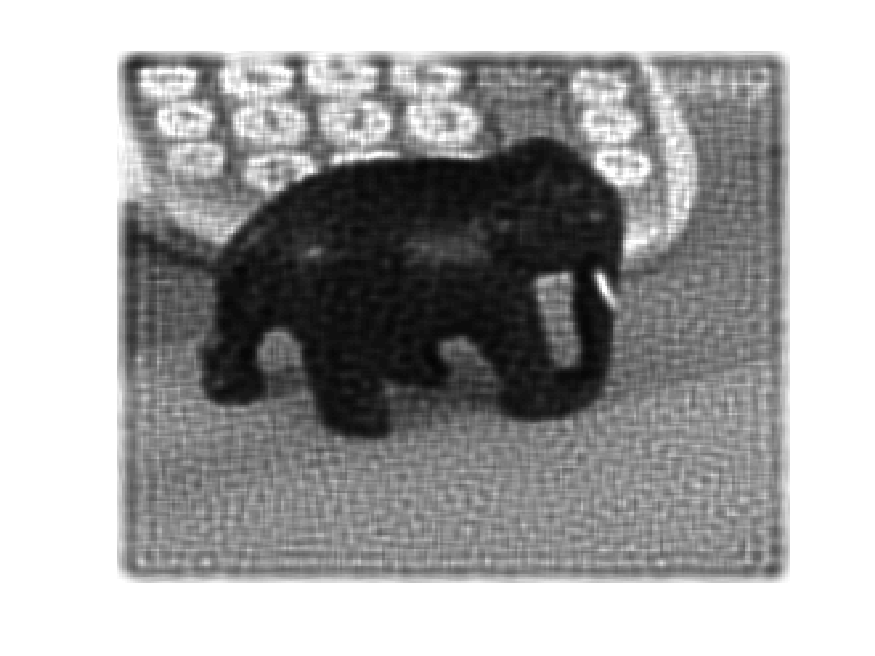}}
\caption{Example 2: Original, blurred noisy image, and various reconstructions, with and without discrete derivative regularization for the tensor-dictionary-based and matrix reconstructions.}
\label{fig:dancerrecon}
\end{figure}

The dictionary used in the reconstruction was constructed from the CalTech face data base (same as in the previous example).  Patch sizes were $16 \times 16$ and we took $s=32$ and used 2000 iterations to obtain each reconstruction (all convergence curves were nearly flat at this point).   The original image in \Cref{fig:a1} was obtained by cropping the
{\sc Matlab} image {\tt clutteredDesk.jpg}.  The tensor dictionary based reconstructions for $\ML$ a discrete gradient operator with $\lambda=10$ is in \Cref{fig:c1}; the tensor-based reconstruction with no regularization is in \Cref{fig:d1}.  The SSIM values of these were $.815$ and $.776$, respectively, showing the insensitivity to $\lambda$ and to additional regularization in general.  The corresponding matrix MRNSD reconstructions with additional regularization (for $\lambda=10$) and without additional regularization.  Without regularization, the borders are white.  The quality depends closely on the value of the regularization parameter, which is problematic.  Moreover, the same or better quality reconstruction can be obtained using
our tensor dictionary based approach without need of choosing a $\lambda$ -- for example, the SSIM of the tensor-based reconstruction in 
(\ref{fig:d1}) was higher than for the matrix case for any value of $\lambda$ that we tried.  

\begin{figure}[h]
\centering
\includegraphics[height=1.75in,width=3.25in]{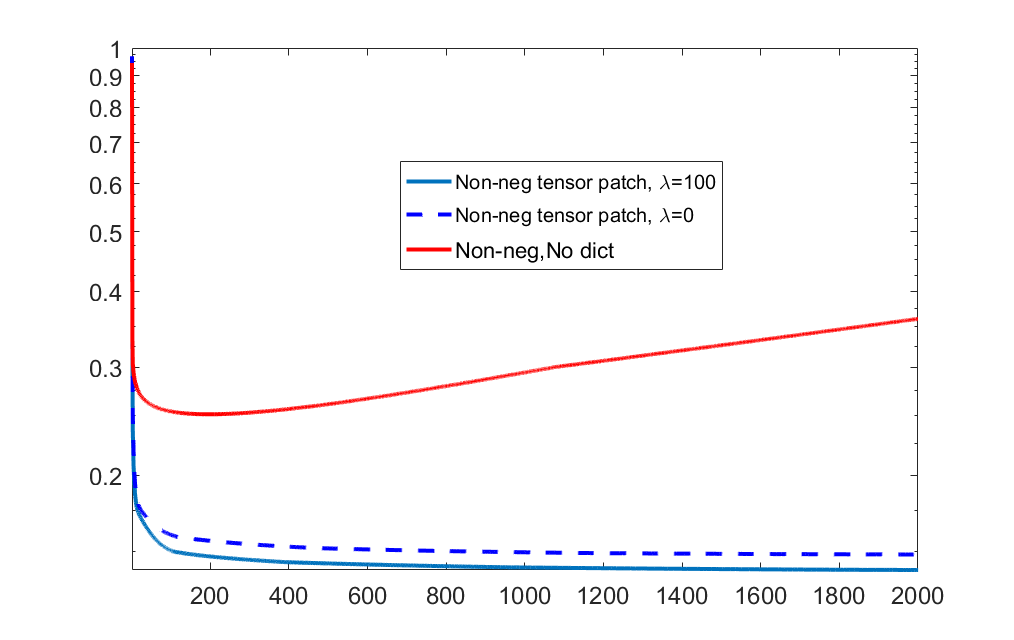}
\caption{\label{fig:a} Example 3:  Convergence behavior.  Horizontal axis is iteration number, vertical is relative error in the iterate against the true image.}
\end{figure}

\begin{figure}[h]
\centering
\subfigure[Blurred, Noisy.]{\label{fig:c} \includegraphics[height=1.25in,width=1.5in]{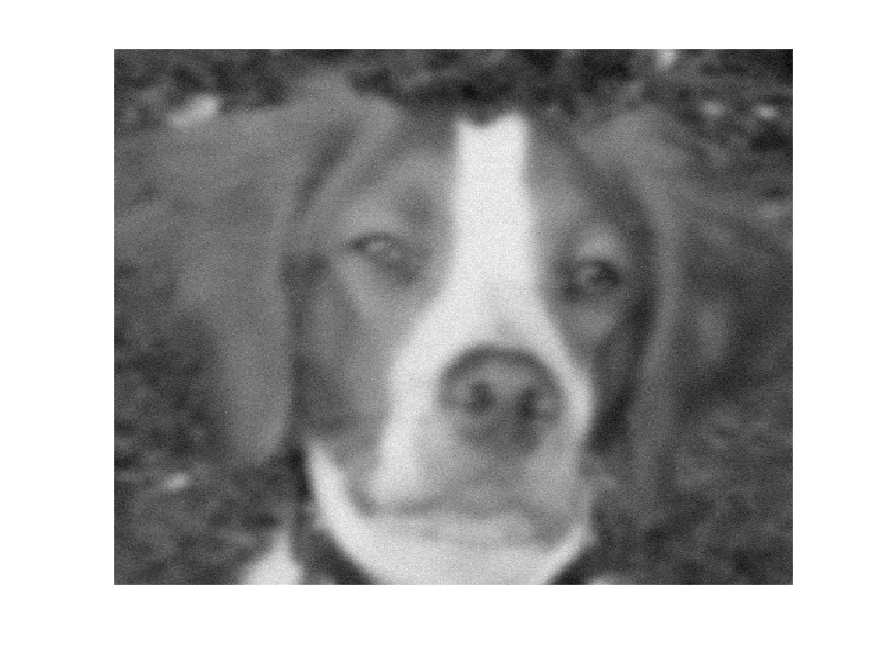}}
\subfigure[Ten $\lambda=100, p,q=16$]{\label{fig:d} \includegraphics[height=1.25in,width=1.5in]{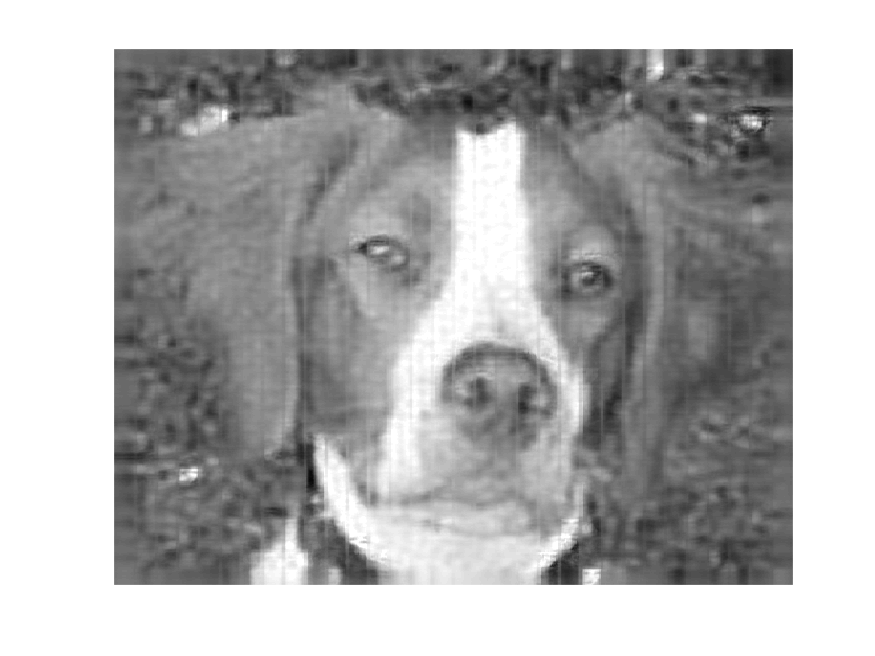}}
\subfigure[Ten  $\lambda = 0, p,q=16$]{\label{fig:e} \includegraphics[height=1.25in,width=1.5in] {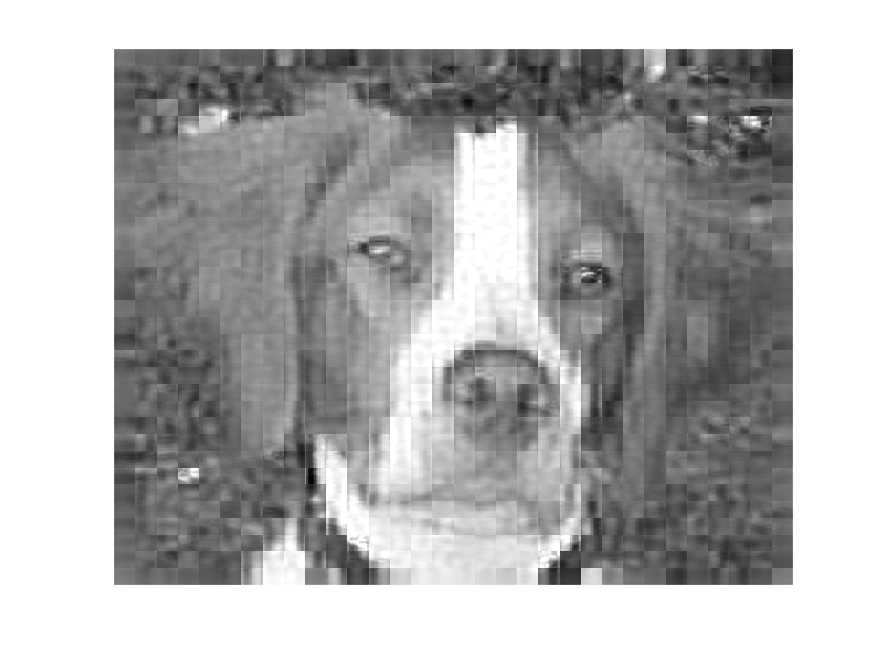}}
\subfigure[Ten  $\lambda = 100,p,q=32$]{\label{fig:f} \includegraphics[height=1.25in,width=1.5in]{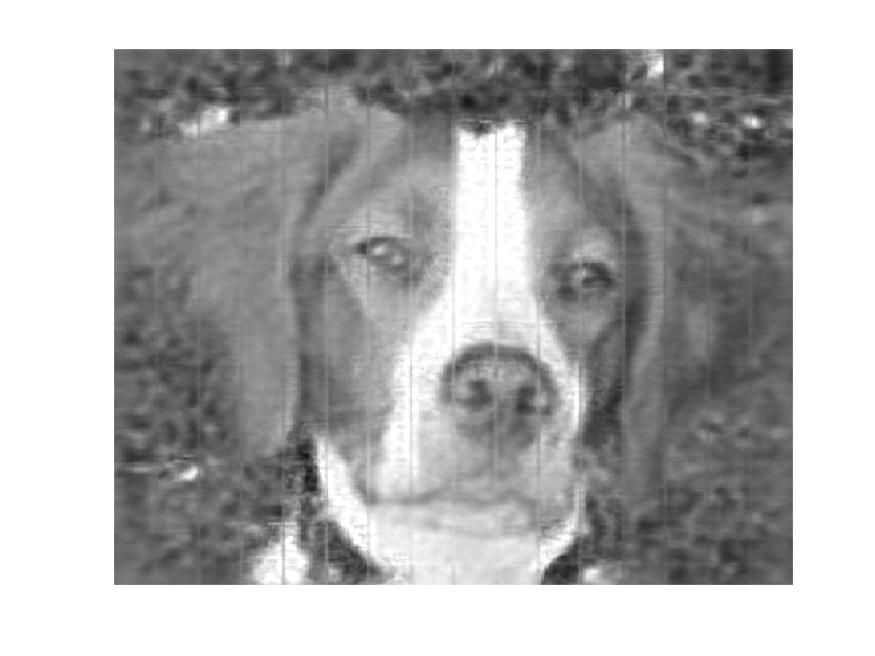}}
\subfigure[Opt. Matrix Recon]{\label{fig:g} \includegraphics[height=1.25in,width=1.5in]{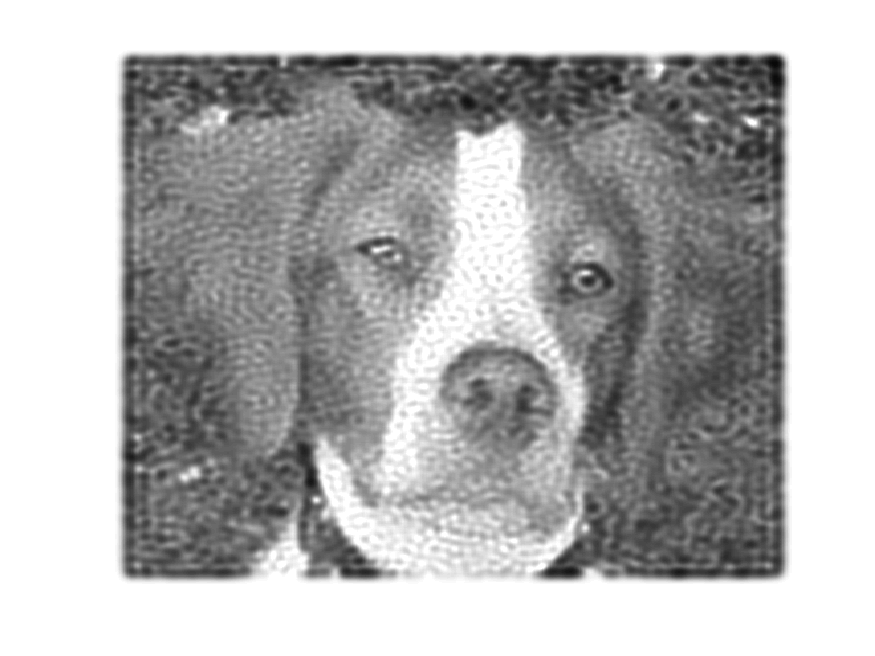}}
\subfigure[Matrix Recon, 2000 its]{\label{fig:h} \includegraphics[height=1.25in,width=1.5in]{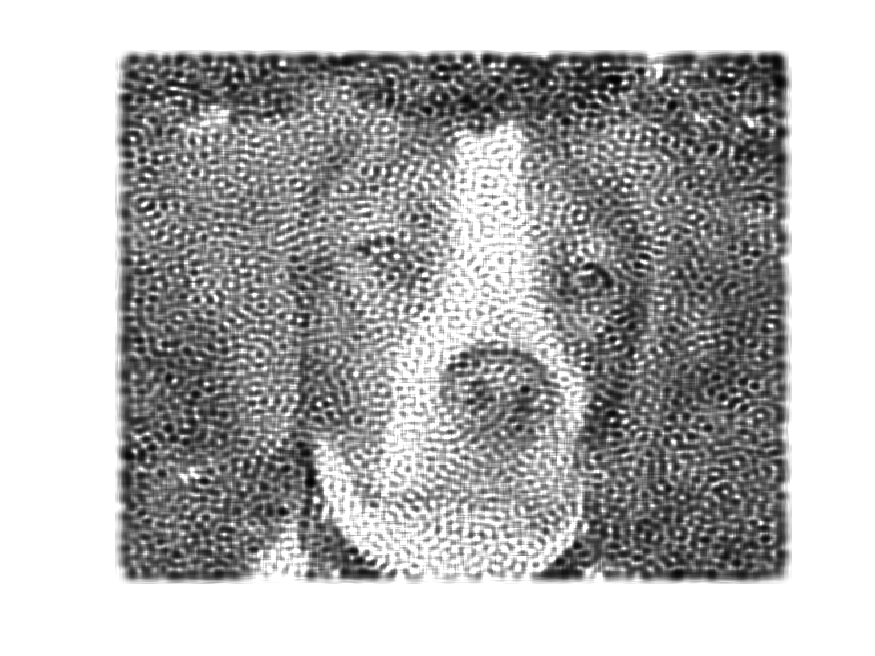}}
\caption{Example 3.  blurred and noisy image, several reconstructions.  In the matrix cases, note the white border in the reconstructions due to the lack of information near the boundary. }
\label{fig:dogrecon}
\end{figure}

In the second illustration, the blurred and noisy image of size $512 \times 512$ is given in \Cref{fig:dogrecon}.   
We use the same dictionary (i.e. learned from human faces) as in the previous illustration, but this time we used a discrete bandwidth of 12, $\sigma=4$, and 5 percent Gaussian noise.  In \Cref{fig:a}, we show the relative errors for using our tensor approach with $\lambda = 100$ and patch-smoothing regularizer, the tensor approach with $\lambda = 0$ (i.e. no smoothing across patches), and the matrix-based MRNSD.  We observe that the behavior is similar for the two tensor classes, with a slight improvement in the error observed when using the patch-regularization term.  We note that semi-convergence behavior is observed in the matrix-based case whereas it is not observed in the tensor cases over the first 2000 iterations.   We show the matrix-based reconstruction at the `optimal' iteration count (198) in subfigure \ref{fig:g}, and the reconstruction after 2000 iterations in subfigure \ref{fig:h}.  Even the optimal reconstruction is qualitatively not as good - clearly, there is a white boundary where it could not be reconstructed, and also there are ringing and fine scale noise artifacts in other areas of the matrix-based image as well.  However, details are recovered using the tensor patch-based dictionary.  

In  \Cref{fig:f} we show a reconstruction using a different dictionary, also constructed from face data, of size $32 \times 64 \times 32$ for $\lambda = 100$ and 2000 iterations.   We see that the quality is very close to the $16 \times 16$ patch dictionary, and there are improvements in some areas of the image but subtle degradataion in others.   As we note in the conclusions, this suggests using a multilevel dictionary approach may improve the situation further.

\section{Summary and Future Work}  \label{sec:future}

In this work, we have shown the utility of learned tensor-patch dictionaries in the context of non-negative image representation, compression and image deblurring applications.  In all cases, once a non-negative tensor patch-dictionary is available, we showed that the problems of compression and deblurring could be formulated in terms of recovering the corresponding non-negative tensor coefficient object.  We gave an MRNSD tensor algorithm for finding the coefficient tensor, and 
described a modification that encourages sparsity in the coefficient tensor.   Notably, this sparsity constraint is applicable whether or not one uses matrices or tensors in the formulation, thereby indicating the proposed approach has broader utility than for the purpose described here.  In the case of deblurring, we showed the tensor representation is particularly effective in mitaging the effects of noise on the solution, especially in the case of underdetermined problems and boundary effects.  

 Importantly, we demonstrated that the class of data on which the dictionary is trained is surprisingly irrelevant in the context of image representation under the tensor-dictionary formulation, as is the resolution of the training data, both in the context of 
image compression and image deblurring. 
We also discussed issues related to patch size, and the trade-offs between sparsity, representability and computation time.    
We showed that a fixed dictionary can do remarkably well on representing images at various resolutions, and even across color channels.  In our deblurring examples, we saw that the tensor dictionaries could mitigate semi-convergence behavior.  We also observed that better representations could be obtained by convex combination of 
deblurred images constructed using dictionaries at different resolutions, which suggests further work is needed to design a multi-level dictionary representation that allows for better local feature description.    Finally, as noted in \cite{Martin:2013kx, Soltani2016}, the t-product generalizes to tensors of order higher than three, so our ideas generalize to higher order.  

Recently, in \cite{LAA}, new tensor-tensor products have been defined which, like the t-product, permit a linear algebraic-type framework.  The tensor-patch dictionary learning and representation approach can therefore be extended to these tensor-tensor products.  Some of the preliminary details are offered in \cite{LizThesis}.   Further investigation into which class of tensor-tensor products provide for the best non-negative dictionaries for use in image compression and representation still needs to be considered, and is the subject of future research.

\appendix

\section{Tensor-based MRNSD derivation}
\label{sec:mrnsdderivation}


Suppose $\T{C} = e^{\T{Z}}$ 
meaning $\T{C}_{ij}^{(k)} = e^{\T{Z}_{ij}^{(k)}}$.  We then compute the 
search direction $\T{S}$ by computing the gradient of \ref{eqn:mrnsd formulation} as follows:
\begin{eqnarray} 
	\T{S} &=& \nabla_{\T{Z}} \left(\frac{1}{2} \| \T{B} - \T{D} * e^{\T{Z}} \|_F^2 \right) \\
		  & = & e^{\T{Z}}\odot (-{\T{D}}^T*(\T{B} - \T{D} * e^{\T{Z}}  )).   \nonumber
	\end{eqnarray}	
	
The search direction $\T{S}$ is exactly the gradient of \cref{eqn:mrnsd formulation} with the addition of a Hadamard product $\odot$ with $\T{C}$.  

To determine the optimal step size, we solve for $\alpha$ as follows.  For notational simplicity, we define the residual tensor $\T{R}$, the gradient tensor ${\T{G}}$, and ${\T{U}}$ as follows:
	\begin{align*}
	\T{R} &= \T{B} - \T{D} * \T{C}\\
	{\T{G}} &= -\T{D}^T * \T{R}\\
	{\T{U}} &= \T{D} * \T{S}
	\end{align*}
Note that ${\T{U}}^T * \T{R} = -\T{S}^T * {\T{G}}$.  We reformulate \cref{eqn:mrnsd formulation} using \cref{defn:fronorm} in terms of $\T{R}$, ${\T{G}}$, and ${\T{U}}$ as follows:
	\begin{align*}
	\frac{1}{2}||\T{B} - \T{D}*(\T{C} - \alpha\cdot \T{S})||_F^2
	&= \frac{1}{2} ||\T{R} + \alpha \cdot {\T{U}} ||_F^2\\
	&= \frac{1}{2} \texttt{trace}[((\T{R} + \alpha \cdot {\T{U}})^T*(\T{R} + \alpha \cdot {\T{U}}))^{(1)}]\\
	&= \frac{1}{2}\texttt{trace}[(\T{R}^T*\T{R} 
		+ 2\alpha\cdot {\T{U}}^T *\T{R} + \alpha^2\cdot {\T{U}}^T*{\T{U}})^{(1)}].
	\end{align*}
Note that typically ${\T{U}}^T*\T{R} \not= \T{R}^T*{\T{U}}$; however, the trace of the first frontal slice is always equal.  We made use of this fact in the last line above.

Now, we solve for $\alpha$ as follows:
	\begin{align*}
	\nabla_\alpha \frac{1}{2}\texttt{trace}[(\T{R}^T*\T{R} 
		+ 2\alpha\cdot {\T{U}}^T *\T{R} + \alpha^2\cdot {\T{U}}^T*{\T{U}})^{(1)}] 
		&= 0\\
	 \texttt{trace}[({\T{U}}^T *\T{R} + \alpha \cdot {\T{U}}^T*{\T{U}})^{(1)}] &= 0
	\end{align*}
Solving for $\alpha$ and rewriting in terms of $\T{D}$, $\T{S}$, and ${\T{G}}$, we get the optimal step size:
	\begin{align*}
	 \alpha &= -\texttt{trace}[({\T{U}}^T *\T{R})^{(1)}]/\texttt{trace}[({\T{U}}^T *{\T{U}})^{(1)}]\\
	 	&=\texttt{trace}[(\T{S}^T *{\T{G}})^{(1)}]/||\T{D} * \T{S} ||_F^2.
	\end{align*}

We add an additional constraint on the $\alpha$ to ensure that we never move too far along the search direction and turn some coefficients $\T{C}$ to negative values.
	\begin{align*}
	\theta &= \texttt{trace}[(\T{S}^T *{\T{G}})^{(1)}]/||\T{D} * \T{S} ||_F^2\\
	\alpha &=  \min\{\theta, \min_{\T{S}_{ij}^{(k)} > 0}({\T{X}}_{ij}^{(k)}/\T{S}_{ij}^{(k)}) \}.
	\end{align*}

\section{Quasiconvexity of MRNSD}
\label{sec:quasiconvex mrnsd}

From Boyd and Vandenberghe's \href{https://web.stanford.edu/~boyd/cvxbook/bv_cvxbook.pdf}{\color{cyan}Convex Optimization}, we have the following definition:

\begin{definition}[Quasiconvex]

A function $f: \Rbb^n \to \Rbb$ is \emph{quasiconvex} if all of its sublevel sets	
	$S_\alpha = \{\V{x} \in \Rbb^n \mid f(\V{x}) \le \alpha\}$ for $\alpha\in \Rbb$ are convex.

\end{definition}

We first expand $\Phi$ as follows:
	\begin{eqnarray*}
	\Phi(\V{z}) & = & \tfrac{1}{2} \| \M{D} e^{\V{z}} - \V{b} \|_F^2  \\
		& = & \tfrac{1}{2} \| \M{D} e^{\V{z}} \|^2 + \tfrac{1}{2} \| \V{b} \|_F^2 - \V{b}^T \M{D}e^{\V{z}}
	\end{eqnarray*}

Suppose for some $\V{x}, \V{y} \in \Rbb^n$ and $\alpha \in \Rbb$, $\V{s}, \V{y} \in S_\alpha$; that is, $\Phi(\V{x}), \Phi(\V{y}) \le \alpha$.  We show that $\theta \V{x} + (1-\theta) \V{y} \in S_\alpha$ for all $\theta \in (0,1)$, hence that $S_\alpha$ is convex and $\Phi$ is quasiconvex.
	\begin{align*}
	\Phi(\theta \V{x} + (1-\theta)\V{y} ) 
		&=
		\frac{1}{2} \| \M{D} e^{\theta \V{x} + (1-\theta)\V{y} }\|_F^2 + \frac{1}{2}\| \V{b} \|_F^2- \V{b}^T \M{D}e^{\theta \V{x} + (1-\theta)\V{y}} \\
				&\le \frac{\theta}{2} \| \M{D} e^{\V{x}} \|_F^2 
				+  \frac{1-\theta}{2}\| \M{D} e^{\V{y}}\|_F^2 + \frac{1}{2}\| \V{b} \|_F^2 
				- \V{b}^T \M{D}e^{\theta \V{x} + (1-\theta)\V{y}} & \text{by convexity}\\
				&\le \theta \alpha + (1-\theta) \alpha- \V{b}^T \M{D} e^{\theta \V{x} + (1-\theta)\V{y}}  
					& \text{by assumption}\\
				&=\alpha- \V{b}^T \M{D}e^{\theta \V{x} + (1-\theta)\V{y} }  
	\end{align*}

For our dictionary-learning problem, we assume $\M{D}$ and $\V{b}$ are non-negative because they are composed of images.  Furthermore, $e^{\V{z}}$ has non-negative components.  Therefore,
	\begin{align*}
	\Phi(\theta \V{x} + (1-\theta)\V{y} ) \le \alpha- \V{b}^T \M{D}e^{\theta\V{x} + (1-\theta)\V{y}}   \le \alpha.
	\end{align*}

Therefore, $\theta \V{x} + (1-\theta)\V{y}  \in S_\alpha$ and $S_\alpha$ is convex.  
Because $\Phi$ is quasiconvex, gradient descent will make progress towards a minimum (i.e., we will not be stuck at a saddle point).

\section*{Acknowledgments}
The authors are extremely grateful to Dr. Sara Soltani for providing us the ADMM code we used to produce the dictionaries. 

\bibliographystyle{siamplain}

\end{document}